\documentclass[11pt, a4paper, logo, copyright]{eai}
\usepackage{shlab}

\usepackage[authoryear, sort&compress, round]{natbib}
\usepackage[]{mdframed}

\usepackage{dblfloatfix}

\usepackage{amsmath,amsfonts,bm}
\usepackage{multirow}
\usepackage{subcaption}
\usepackage{wrapfig}

\def\eqref#1{equation~\ref{#1}}

\def\1{\bm{1}}

\DeclareMathAlphabet{\mathsfit}{\encodingdefault}{\sfdefault}{m}{sl}
\SetMathAlphabet{\mathsfit}{bold}{\encodingdefault}{\sfdefault}{bx}{n}

\usepackage{listings}
\usepackage{hyperref}
\usepackage{url}
\usepackage{graphicx}
\usepackage{amsmath} %
\usepackage{mathrsfs} %
\usepackage{etoolbox}
\usepackage{cleveref}
\usepackage{tcolorbox}
\usepackage{colortbl}
\usepackage{booktabs}       %
\usepackage{amsfonts}       %
\usepackage{nicefrac}       %
\usepackage{microtype}      %
\usepackage{caption}
\usepackage{subcaption}
\usepackage{algorithm}
\usepackage{algorithmic}
\usepackage{multirow}
\usepackage{lipsum}
\usepackage{booktabs} %
\usepackage{enumitem}%

\usepackage{makecell}
\usepackage{adjustbox}

\setlist[itemize]{noitemsep, topsep=0pt}
\usepackage{enumitem,kantlipsum}

\usepackage[normalem]{ulem}
\useunder{\uline}{\ul}{}

\newlength\savewidth

\definecolor{baselinecolor}{HTML}{d6eaf8}

\definecolor{mygray}{gray}{0.4}

\AtBeginEnvironment{tcolorbox}{\tiny}

\newcount\Comments  %
\usepackage{color}
\definecolor{darkred}{rgb}{0.9,0,0}
\definecolor{darkgreen}{rgb}{0,0.5,0}
\definecolor{darkblue}{rgb}{0,0,0.7}
\definecolor{purple}{rgb}{.6, 0,.6}
\definecolor{orange}{rgb}{1.0,0.64,0}
\newcommand{\kibitz}[2]{\ifnum\Comments=1\textcolor{#1}{#2}\fi}

\title{A Vision-Language-Action-Critic Model \\for Robotic Real-World Reinforcement Learning}
\correspondingauthor{$^*$ equal contributions, $^\dagger$ corresponding to (pangjiangmiao@pjlab.org.cn). Contributions and emails of all authors in \Cref{sec:authors}.}

\author[*,1]{Shaopeng Zhai}
\author[*,1]{Qi Zhang}
\author[*,1]{Tianyi Zhang}
\author[*,1]{Fuxian Huang}
\author[*,1]{Haoran Zhang}
\author[*,1]{Ming Zhou}
\author[1]{Shengzhe Zhang}
\author[1]{Litao Liu}
\author[1]{Sixu Lin}
\author[$\dagger$,1]{Jiangmiao Pang}

\affil[1]{Shanghai AI Lab}

\begin{document}
\begin{abstract}
Robotic real‑world reinforcement learning (RL) with vision‑language‑action (VLA) models is bottlenecked by sparse, handcrafted rewards and inefficient exploration. 
We introduce VLAC, a general process reward model built upon InternVL and trained on large scale heterogeneous datasets.
Given pairwise observations and a language goal, it outputs dense progress delta and done signal, eliminating task‑specific reward engineering, and supports one‑shot in‑context transfer to unseen tasks and environments. 
VLAC is trained on vision–language datasets to strengthen perception, dialogic and reasoning capabilities, together with robot and human trajectories data that ground action generation and progress estimation, and additionally strengthened to reject irrelevant prompts as well as detect regression or stagnation by constructing large numbers of negative and semantically mismatched samples.
With prompt control, a single VLAC model alternately generating reward and action tokens, unifying critic and policy.
Deployed inside an asynchronous real‑world RL loop, we layer a graded human‑in‑the‑loop protocol (offline demonstration replay, return and explore, human guided explore) that accelerates exploration and stabilizes early learning. 
Across four distinct real‑world manipulation tasks, VLAC lifts success rates from about 30\% to about 90\% within 200 real-world interaction episodes; incorporating human‑in‑the‑loop interventions yields a further 50\% improvement in sample efficiency and achieves up to 100\% final success.

\links{
  \link{code}{Code:VLAC}{https://github.com/InternRobotics/VLAC}, 
  \link{model}{Model:VLAC}{https://huggingface.co/InternRobotics/VLAC}, 
  \link{homepage}{Homepage \& Interactive-demo}{https://vlac.intern-ai.org.cn/}, 
}
\vspace{0.5cm}

\end{abstract}

\maketitle

\begin{quote}
\textit{"Intelligence is determined by the dynamics of interaction with the world."} \\
{\raggedleft — Rodney A. Brooks, 1991\par}
\end{quote}

\section{Introduction}
\hspace{1.5em}With the rapid development of Vision-Language-Action (VLA) models, the intelligence of robotic perception and manipulation capabilities has greatly improved, leading to impressive performance in autonomously completing general tasks. Current VLA models are primarily trained through imitation learning, which requires vast amounts of data and demands substantial human and material resources for large-scale data collection~\cite{wang2025genie,lin2024data,team2025gemini,deng2025graspvla,bjorck2025gr00t}. However, collecting human expert trajectories is not only costly and time-consuming, but most data collection efforts focus on laboratory-customized scenes and tasks, with inconsistent quality. Consequently, significant barriers remain for robots to perform effectively in real-world scenarios, particularly concerning data diversity, cross-scene generalization, and robustness.

\begin{figure}[t!]
    \centering
    \includegraphics[width=.95\linewidth]{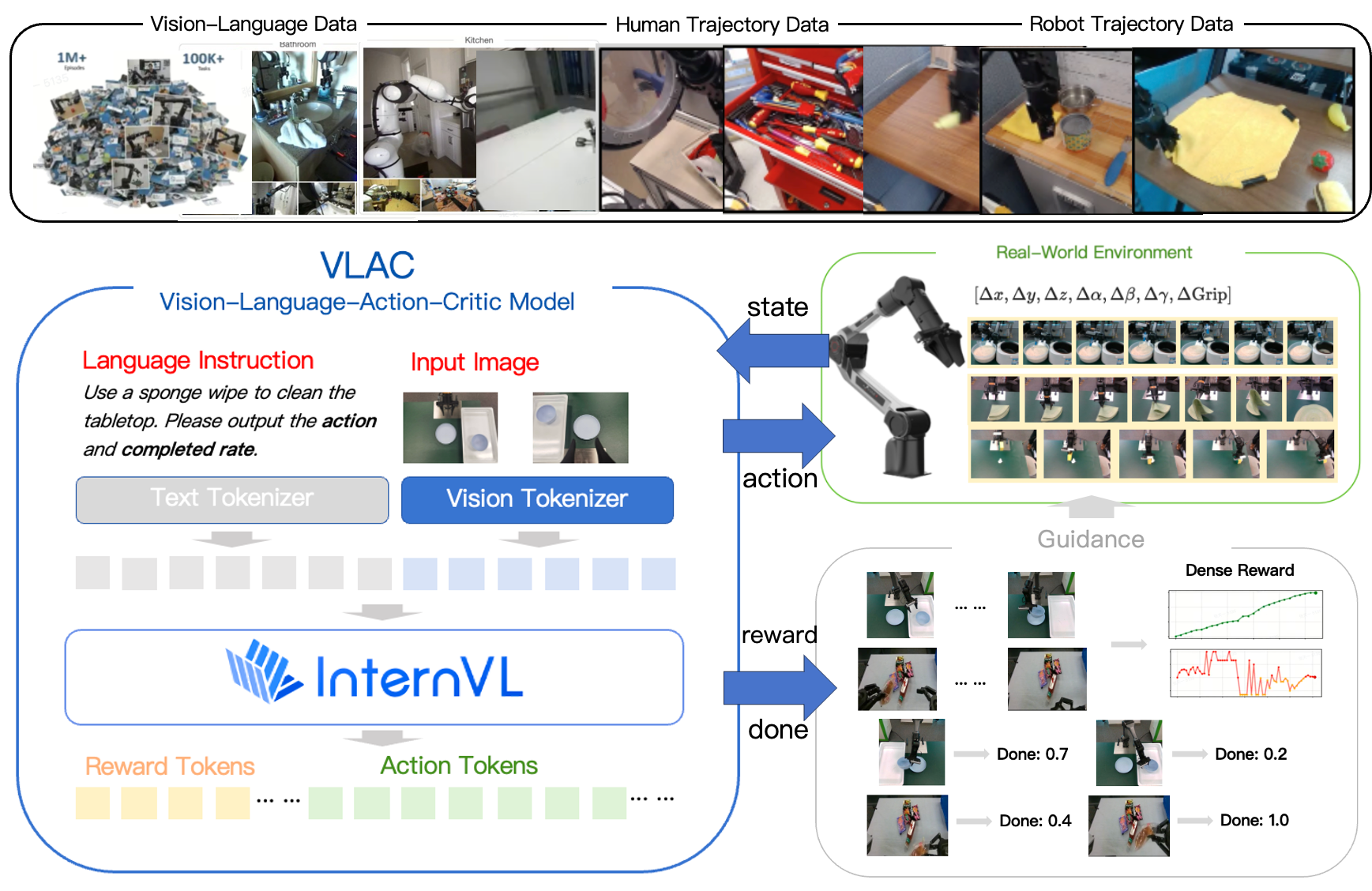}
    \caption{\textbf{Overview}. Pretrained on multi-source data, VLAC provides dense progress rewards for real-world RL while also serving as a policy to output actions, integrating into real-world RL loops to enable self-improvement in manipulation.} %
    \label{fig:teaser}
\end{figure} 

To address this limitation, which is often insufficient to cover the dynamic and continuously changing tasks and environments encountered in the real world, and to avoid the challenges associated with sim-to-real transfer, many studies have explored RL conducted directly in the real world. In this paradigm, robots autonomously learn from both successful and failed trajectories generated through real-world interactions, enables autonomous exploration and continuous improvement of success rates from real-world feedback.
A straight forward way for improving real‑world RL efficiency is to provide dense progress reward. 
However, real‑world RL reward design still encounters persistent obstacles. Many methods rely on non‑general, task‑specific shaping engineered separately for each scenario ~\cite{mendonca2024continuously,herzog2023deep,mendonca2023alan,kumar2024practice,xu2022dexterous}. Frameworks advertised as reusable often require additional, task‑dependent data collection to train reward surrogates or termination (done) classifiers ~\cite{luo2024precise,hu2023reboot}. The resulting rewards typically appear only when the task is nearly complete, leaving most intermediate progress unscored. Although some approaches introduce denser or “universal” progress signals, their ability to generalize across novel tasks, objects, or goal language remains limited ~\cite{ma2023liv,ma2022vip}. Thus, intermediate feedback is still unreliable and weakly transferable, impeding sample efficiency. We target this gap by producing dense progress reward with stronger cross‑task generalization.

To provide reliable and generalizable reward signal, we propose a Vision-Language-Action-Critic model, VLAC, which builds upon InternVL, a state-of-the-art multi-modal model, and unifies the roles of ``actor'' and ``critic'' within a single autoregressive architecture, capable of zero-shot and in-context task progress prediction and action generation.
VLAC takes as input a pair of image observations together with a task description and outputs a signed progress delta indicating how much the second state advances (or regresses) the task relative to the first, as the reward.
It is enabled by training on more than 4,000 hours of language annotated manipulation data where temporal ordering yields the progress labels, and we construct additional data to strengthen negative reward assignment and improve in context learning capability. Auxiliary perception tasks such as detection, lightweight segmentation, grounding, and coarse 3D or contact cues enhance the base representation.
Experiments show that VLAC separates positive and negative progress with sufficient fidelity to yield a reliable reward signal. And we find that enhancing the critic component leads to a improvement in downstream action generation.

Building upon the VLAC model, we developed a real-world RL framework in which the VLAC model can interact directly with the real world and self-improve. Through adaptive interaction, the VLAC collects both successful and failed trajectories, receives reward signals, and determines task completion. We evaluated our framework on four diverse real-world manipulation tasks, where the ``actor'' demonstrated the ability to autonomously improve its success rate from approximately 30\% to 90\% within 200 episodes.
During the experiments, we observed that the VLA's prior capabilities play a critical role in enabling efficient exploration. 
To further enhance exploration efficiency, we introduced a human-in-the-loop mechanism. This mechanism allows humans to assist the model in refining the manipulation process, facilitating more effective learning. Human intervention is more of an art than a science, and we applied three levels of human-in-the-loop intervention across all tasks, resulting in an approximate 50\% improvement in exploration efficiency and achieves up to 100\%  success rate.

\section{Related Work}
\label{sec:related}

\hspace{1.5em}\textbf{Real-world RL} has advanced rapidly across multiple robotic subfields, including locomotion\cite{smith2024grow,smith2023demonstrating} and dexterous manipulation\cite{pmlr-v229-hu23a}. A substantial portion of “online” approaches, however, function essentially as self‑imitation\cite{kumar2024practice} or hindsight relabeling\cite{zhou2024autonomous}: they reinforce previously successful behaviors without exploiting failed (negative) attempts, which limits overall data efficiency in real‑world settings. Focusing on manipulation, one prominent line of work applies real‑world RL to small (non‑VLA) policy models. Because such models lack strong prior ability, researchers commonly adopt human‑in‑the‑loop procedures\cite{luo2024hilserl,li2025reinforcement,kang2025forget,zhou2024efficient}: a small set of expert demonstrations is first collected, then offline‑to‑online reinforcement learning repeatedly leverages these demonstrations to improve sample efficiency during interaction. Although effective, this imposes considerable engineering overhead: for each new task practitioners typically must (1) collect and curate task‑specific data, (2)hand‑design or shape reward signals, and (3) implement bespoke task‑completion (done) detectors. More recently, another direction explores real‑world RL atop large Vision‑Language‑Action (VLA) models\cite{chen2025conrft,lv2025flow,park2025flow}. Their strong pretrained multimodal priors markedly accelerate early exploration and raise the probability of discovering successful behaviors. Nevertheless, heterogeneity across existing VLA architectures—spanning perception–language fusion, action representation and formatting, and progress/value evaluation interfaces—has led to a diverse and fragmented set of RL integration strategies, with no clear convergence of design patterns so far.

\textbf{General reward models}, In many real‑world RL settings, tasks are initially formulated with only a terminal success/failure or extremely sparse segmented signal, exacerbating both exploration depth and credit assignment challenges\cite{luo2024serl,chen2025conrft,zhou2024efficient}. A direct route to improved sample efficiency is to construct dense and temporally consistent progress signals. Because, under the Markov assumption, rewards can be treated as functions of the current observation plus a goal specification (without explicit action dependency), large-scale action‑free internet/embodied video and pretrained multimodal models (VLMs or text–image alignment models) have been leveraged to build transferable value/progress functions. Existing approaches fall broadly into four categories: (1) Prompt‑based VLM scoring\cite{wang2024,10610873}: prompting a general VLM to estimate completion for a single state or to compare pairs/adjacent states; (2) Semantic embedding distance (e.g., CLIP-style)\cite{xiong2024adaptive,10161016}: mapping states and goal descriptions into a shared embedding space and using similarity as instantaneous reward; (3) Goal state synthesis/editing\cite{zhou2024autonomous}: employing image editing or conditional generation models to produce an “ideal” target image from the current state and goal description, then transforming the difference into a reward; (4) Learned progress embeddings (implicit time‑contrastive / temporal‑difference)\cite{zhangefficient,ma2022vip,ma2023liv,biza2024robot}: enforcing temporal ordering, local smoothness, and contrastive constraints on embodied or human demonstration videos so that embeddings become (piecewise) rank‑consistent with task progression; distances between current and goal embeddings then implicitly define value/reward for arbitrary goal images or goal text.
While directly querying generic VLMs provides rapid, task‑agnostic scoring interfaces, such “external” reward sources suffer from limited spatial/geometric precision, sensitivity to viewpoint/occlusion/lighting shifts, temporal instability (non‑monotonic framewise scores), and weak discrimination of failed trajectories—collectively inflating advantage variance and destabilizing policy updates. In contrast, time‑contrastive or TD‑style progress representation learning can yield smoother reward signal in an episode, but lacks generalization
capabilities\cite{ma2024vision}. Key open questions includes: (a) fine‑grained yet low‑noise progress estimation, (b) strong discriminability for failures and deviations, and (c) tight integration with multimodal perception and action generation in a single end‑to‑end architecture.

\textbf{RL post‑training for VLAs}, especially for vision–language models: most systems employ a single autoregressive architecuture, so policy gradient methods—typically PPO\cite{schulman2017proximal} variants or REINFORCE objectives (e.g., GRPO\cite{shao2024deepseekmath})—operate directly on token log‑probabilities. In Vision - Language - Action (VLA) settings, in contrast, the architectures are heterogeneous: some produce discrete action tokens or structured textual templates\cite{kim24openvla,pertsch2025fast,zhai2024buildingopenendedembodiedagent}, others produce continuous action\cite{kim2025fine,black2024pi_0,chi2023diffusion}, preventing a uniform transplantation of token‑centric RL recipes.


For discrete autoregressive action heads, token‑level PPO remains a natural choice. When actions are produced via diffusion or flow‑matching (continuous iterative decoding), Two broad patterns appear. (1) Value‑guided filtering or weighting (imitation‑flavored): rejection sampling\cite{he2024aligniql} or importance weighting emphasizes high‑Q/advantage samples\cite{zhangenergy}, effectively steering the generative distribution toward higher‑return behaviors. (2) Offline‑regularized RL: maximize value with an auxiliary BC regularization term that keeps the learned policy close to the expert (or prior) distribution, thereby reducing distributional shift and extrapolation errors~\cite{lv2025flow,park2025flow}. Because diffusion and flow matching rely on multi‑step denoising or ODE integration, naïve end‑to‑end backpropagation of RL signals through every iteration creates long gradient chains (BPTT‑like) with instability and computational overhead\cite{lv2025flow}. And solution include: aligning the gradients of policy directly to the Q during denoising process\cite{psenka2024qsm}; truncating gradients to the final K denoising/integration steps\cite{dppo2024}; collapsing the multi-step process into a single-step policy via distillation\cite{park2025flow,li2025reinforcement} or mean-flow\cite{geng2025mean}.

Strong pretrained (zero/few‑shot) capability is central in real‑world VLA RL, and Offline‑to‑Online strategies preload buffers with human or demonstration data to reduce early exploration cost\cite{luo2024hilserl}. These approaches borrow from offline RL regularization: addressing value underestimation\cite{zhou2024efficient}, and coupling Q maximization with a BC regularizer to limit distributional drift\cite{kang2025forget,chen2025conrft,dingconsistency}. They typically emphasize maintaining proximity to initial human data while enabling gradual policy improvement through selective reuse of stored trajectories.

\section{VLAC model and Real-World RL Framework}

\hspace{1.5em}We propose a unified VLAC model that leverages multi-source data to enhance both task progress critic and action policy capabilities. The VLAC model can accurately predict task progress and generate actions, demonstrating strong generalization to unseen environments and tasks. Based on VLAC, we further design a real-world reinforcement learning framework and incorporate a human-in-the-loop framework to improve efficiency, enabling self-improvement in real-world settings. In the following sections, we provide detailed introductions to Vision-Language-Action-Critic Model and Real-world RL with VLAC.


\begin{figure}[t!]
    \centering
    \includegraphics[width=.95\linewidth]{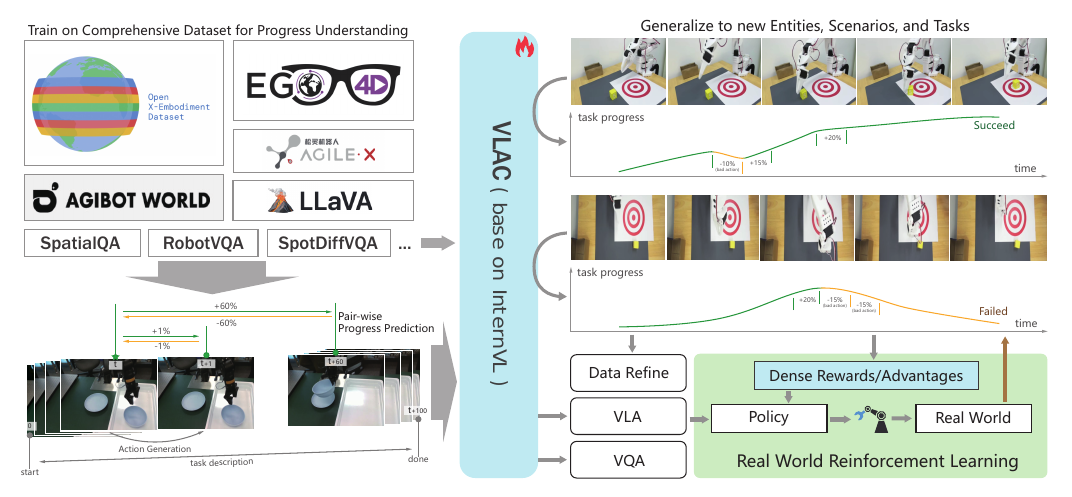}
    \caption{The VLAC model is trained on a combination of comprehensive public robotic manipulation datasets, human demonstration data, self-collected manipulation data, and various image understanding datasets. Video data is processed into pair-wise samples to learn the different task progress between two frames, supplemented with task descriptions and task completion evaluation to enable task progress understanding and action generation, as illustrated in the bottom-left corner. As shown in the diagram on the right, the model demonstrates strong generalization capabilities to new robots, scenarios, and tasks not covered in the training dataset. It can predict task progress and distinguish failure action or trajectory, providing dense reward feedback for real-world RL and offering guidance for data refinement. Additionally, the model can directly perform manipulation tasks, exhibiting zero-shot capabilities to handle different scenarios.} %
    \label{fig:VLAC}
\end{figure}

\subsection{Vision-Language-Action-Critic Model}
\hspace{1.5em}We constructed a Vision-Language-Action-Critic Model based on pair-wise progress understanding, enabling joint training with both action robotics data and non-action human data, as shown in Figure~\ref{fig:VLAC}. This model jointly realizes action generation and delta task progress generation, featuring in-context learning and cross-scenario, cross-task generalization capabilities. 
The pairwise progress module takes as input two images corresponding to arbitrary intermediate task states; it is agnostic to data collection strategy and to segment starting points, thereby improving robustness and broad applicability. Its output is a delta progress value: a positive value indicates that the second image reflects a more advanced stage of task completion (a negative value indicates the opposite). This value naturally serves as a temporal-difference (TD) reward for reinforcement learning and fine-grained dataset quality estimation.
Our VLAC model is trained on more than 3000 hours of human data, 1200 hours of comprehensive public robotic manipulation data, and over 15 hours of self-collected manipulation data, providing action control, robust task progress rate assessment, and task completion verification.


\subsubsection{VLA\textcolor{red}{\textbf{C}} critic learning}
\hspace{1.5em}Humans often encounter limitations in their initial capabilities when faced with new environments and new tasks. However, their ability to understand the progression of the task is exhibited in a higher generalizability than their ability to execute tasks. This ability enables continuous assessment of the task progress across various processes, thereby optimizing one's actions and achieving sustained improvement. Rich general knowledge thus endows the understanding of task progress with strong generality, which can facilitate capabilities when faced with different environments and tasks. Inspired by this, for robot manipulation models to be deployed in the complexity of the real world and execute general tasks, they must not only understand language and visual information, but also understand task progress, which can indicate changes in task completion status across different processes. We design a pair-wise task progress understanding method that is unaffected by the initial point of the task. This method integrates general human data without action information and robot data annotated with actions to robustly estimate fine-grained variations in task processes.

We formalize the task process as a video segment with description $V = (O, l_\text{task})$, which consists of a basic RGB image sequence $O = (o_1, \ldots, o_T)$ and a textual task goal description $l_\text{task}$. Each training trajectory in our dataset is a successful and efficient execution of the annotated task. To understand the variations of the task process, we assume that task progress is positively correlated with time; as time increases, the task progresses. Thus, pair-wise task progress understanding can be formalized as
\begin{equation}
c_{i,i+\Delta t} = \mathrm{VLAC}(o_i, o_{i+\Delta t}; l_\text{task}),
\end{equation}
where $\Delta t$ denotes the time difference between two frames, and $c_{i,i+\Delta t}$ represents the degree to which the task progresses in $o_{i+\Delta t}$ advances the task relative to $o_i$. Specifically, $\Delta t \in [-i+1, T-i] \cap \mathbb{Z}$, which allows us to focus on both fine-grained single-step changes and long-term task progress, mitigating noise from minor variations while naturally constructing balanced negative samples. During training, $c_{i,i+\Delta t}$ is annotated according to the natural temporal order as $c_{i,i+\Delta t} = \Delta t / (T-i)$, representing the percentage of task progress from $o_i$ to $o_{i+\Delta t}$. This progress estimation formulation treats any sampled sub‑segment of a trajectory as a self‑contained unit: it measures only the relative advancement between the two frames within that local window and is agnostic to the trajectory’s global start point or data collection strategy, thereby improving generality and robustness, as illustrated in the lower left corner of Figure~\ref{fig:VLAC}.

To further enhance the semantic understanding of the task process, we construct a task description estimation objective: 
\begin{equation}
l_\text{task} = \mathrm{VLAC}(o_{i_\text{start}}, o_{i_\text{end}}),
\end{equation}
where $o_{i_\text{start}} \in [0, 0.3T] \cap \mathbb{Z}$ and $o_{i_\text{end}} \in [0.8T, T] \cap \mathbb{Z}$. By generating task descriptions from the initial and final frames, the task description becomes not only an input condition but also an output target, thereby improving the joint understanding of vision and language.

To enhance understanding of task completion, we design a task completion judgment task: 
\begin{equation}
l_\text{done} = \mathrm{VLAC}(o_i; l_\text{task}),
\end{equation}
where if $i < 0.8T$, $l_\text{done} = 0$ indicates the task is not yet completed, and if $i > 0.95T$, $l_\text{done} = 1$ indicates the task is completed. Considering the diversity of data and collection strategies, it is difficult to accurately determine the exact completion point; thus, for $0.8T \leq i \leq 0.95T$, no training label is made to ensure label accuracy. By learning to judge task completion, the model’s understanding of completion conditions is enhanced, providing auxiliary signals for task completion in real-world RL.

We use different prompts to distinguish different tasks. Furthermore, to improve the task progress understanding, we design four data construction strategies:

1. Pair-wise Image Difference Filtering:  
   We assume task progress is positively correlated with time, which generally holds but may be violated in noisy data or segments with minimal change (e.g., static scenes). To mitigate the impact of such noise, we set the interval between $i$ and $i+1$ to approximately 0.2s during data construction and compare the pixel difference between the two frames. If $\mathrm{Diff}(o_i, o_{i+\Delta t}) < \sigma$, then set $c_{i,i+\Delta t} = 0$, indicating the two frames are in the same progress. In our experiments, we set $\sigma = 1\%$, enabling the model to focus on significant changes and improving the robustness of task progress understanding.

2. Pair-wise Progress Understanding with Joint Sampling:
   Inspired by contrastive learning, to ensure data balance and symmetry between forward and reverse processes, for each sampled pair $(o_i, o_{i+\Delta t})$, we construct four related data samples as a mini group within a batch:
   $$
   \begin{cases}
   (o_i, o_{i+1}) \\
   (o_{i+1}, o_i) \\
   (o_i, o_{i+\Delta t}) \\
   (o_{i+\Delta_t}, o_i)
   \end{cases}
   $$
   This covers both forward and backward processes, as well as fine-grained and global understanding, as shown in the lower left of Figure~\ref{fig:VLAC}.

3. Task Completion Judgment Joint Sampling: 
   For data balance in the task completion judgment task, each time we sample a pair of data: one from a completed state and one from an incomplete state within the same trajectory.

4. Cross-sampling of Task Descriptions and Image Sequences:  
   To enhance the model’s ability to distinguish whether a process matches the task description, we sample, with a 5\% probability during pair-wise data sampling, a task description $l_\text{task}$ that does not belong to the current trajectory, setting $c_{i,i+\Delta t} = 0$. This method aims to improve the alignment of semantic and progress understanding of the model.

Cross-scene and cross-task transferability remains a key challenge for embodied intelligence models on the path to generalization. When humans adapt to new environments and tasks, their initial capabilities may also be limited; however, having reference examples can significantly improve both initial performance and learning efficiency. Inspired by this, and to improve the cross-scene and cross-task transferability of VLAC, we further enhance progress understanding with in-context learning, enabling effective learning from a single reference example. Specifically, in-context progress understanding can be formalized as 
\begin{equation}
c_{i,i+\Delta t} = \mathrm{VLAC}(o_i, o_{i+\Delta t}; l_\text{task}, O_\text{ref}, o_0),
\end{equation}
where $O_\text{ref}$ is the reference process, which may be provided by a robot demonstration or a human demonstration, offering guidance on both scene and task logic. $o_0$ is the starting point of the current trajectory and can be optionally included as input, enhancing the model’s ability to align with the reference process and enabling inference of the absolute progress of $o_i$ and $o_{i+\Delta t}$.

These tasks we construct do not require action information, thereby avoiding the issue of inconsistent action spaces across entities. This design also allows our approach to apply to both human and robot data, enabling the use of large-scale, diverse human data to significantly improve model generalization and alleviate the scarcity of robot data in the real world. Moreover, in-context learning endows the model with rapid transfer capabilities and further enhances its generalization.

\subsubsection{VL\textcolor{red}{\textbf{A}}C action learning}
\hspace{1.5em}Based on general task process understanding, we further construct an action generation task in the semantic space to achieve multi-task control of a robotic arm. Since the general generation capabilities of pretrained multimodal models are mainly in the semantic space, and task understanding is also generated in the semantic space, we fully leverage this knowledge and the strong semantic representations of pretrained models by representing actions as numbers and generating them in the semantic space by the autoregressive approach.

To further improve spatial reasoning performance, the action is represented as the delta End-Effector (eef) pose, which is a general spatial three-dimensional representation while remaining independent of embodied entities:  
$$
a_{i} = \mathrm{VLAC}(o^0_i, \ldots, o^k_i; s_i; l_\text{task}; \mathrm{history}_{i-1,i-t_h}),
$$  
where $o^k_i$ denotes the $k$-th viewpoint at the $i$-th step, $s_i$ represents the state of the robotic arm, and $a_{i}$ is the action to be executed at the $i$-th step, represented as a string of numbers. $\mathrm{history}_{i-1,i-t_h}$ is the generated action history from step $i-1$ to $i-t_h$.

With this formulation, VLA exhibits strong semantic and scene generalization capabilities. Additionally, the generated actions can be sampled with diversity within a reasonable range, which is beneficial for exploration and improvement in reinforcement learning.

\subsubsection{\textcolor{red}{\textbf{V}}\textcolor{red}{\textbf{L}}AC vision-language perception learning and data mixture}
\hspace{1.5em}To enhance the model's multimodal understanding capabilities, we incorporate a series of publicly available VQA datasets, focusing on four aspects: general conversational ability, robotic understanding, spatial reasoning, and pair-wise image difference distinction. The details of the datasets are shown in Appendix.\ref{sec:dataset}.
By leveraging these datasets, we aim to comprehensively improve the model’s multimodal reasoning and understanding capabilities across diverse application domains.


For all three tasks above, we collected and processed data from various sources, including human demonstration data, multiple types of robotic arm data, and VQA datasets, i.e., Ego4D HOD~\cite{pei2025modeling}, AGIBOT~\cite{bu2025agibot}, Bridge~\cite{walke2023bridgedata}, Droid~\cite{khazatsky2024droid}, FMB~\cite{luo2025fmb}, RoboSet~\cite{bharadhwaj2024roboagent}, Self Collected dataset, Llava~\cite{liu2023visual}, SpatialQA~\cite{chen2024spatialvlm}, RobotVQA~\cite{sermanet2024robovqa}, Spot the diff~\cite{jhamtani2018learning}, InstructPix2Pix~\cite{brooks2023instructpix2pix}. In addition, we collected a small portion of our data specifically for fine-tuning action representations on our robotic arm. The details of the datasets and their combinations are shown in Table ~\ref{tab:dataset}. In total, we sampled 40 million data points (some of which include multi-turn dialogues) for training.

\subsection{Real-world RL with VLAC}
\subsubsection{Infrastructure of RL}
\hspace{1.5em}We build upon a single-controller architecture, differing from existing RLHF frameworks by focusing on the robot-centric design for real-world reinforcement learning. Observations sent by each robot must be processed promptly to generate actions; otherwise, delays in real-world execution would waste valuable time as the robotic arm waits. To address this, we implemented a dynamic inference server allocation mechanism, ensuring that all robot observations are assigned to idle VLA replicas for inference within 0.1 seconds. Although this approach reduces GPU utilization and results in smaller inference batches for some VLA replicas, it significantly minimizes latency during real-world robot operations.

It is important to note that the robots execute tasks asynchronously, meaning that uploading observations and executing actions can occur independently. Robots do not need to wait for the completion of an action before uploading new observations, which reduces waiting time. However, this asynchronous process means that the generated actions may not correspond to the exact observation at the moment of action generation. To address this, during VLA training, action timestamps are adjusted to lag behind observation timestamps by a duration determined by the VLA's inference time. By coordinating the adjusted action timing with the robot’s motion speed, the next action arrives precisely when the previous action finishes execution, ensuring smooth and continuous robot movements.

The VLA model operates on GPU servers and communicates with robots via ZeroMQ (zmq). The framework is built using Ray and comprises several independent components: inference workers for prediction, trainers for model training, data servers for storage and distribution, log servers for logging aggregation, and rollout workers for handling robot communication and information collection. For inference, both vllm and torch can be used. While vllm provides faster inference, we observed significant discrepancies between vllm and torch results. Under identical neural network parameters and samples, the importance ratios for the same action generated by vllm and torch fluctuate between 0.4 and 1.8. This discrepancy frequently triggers the clipping mechanism in PPO, rendering approximately 60\% of the data unusable. Therefore, when using vllm, despite its faster inference speed, we must recompute the action probabilities with torch during training instead of directly reusing the probabilities obtained during inference.
\subsubsection{PPO based policy optimization}
\hspace{1.5em}During rollout, the inference worker checks whether the task has been done with the VLAC model, using prompt "<image> The 1 means yes, the 0 means no. Check if the robot has completed its task: <task>", and the reward is extracted using prompt "Image-1: <image> Image-2: <image> Compare two images and evaluate whether the second image is closer to achieving task objectives compared to the first image. + score means the second image is closer, - score means the first image is closer Response the relative progressing of target task follow <score>. The target task is: {task} <score>". 
Then Proximal Policy Optimization (PPO) is adopted to improve the policy model, PPO introduces a stable and efficient approach to optimizing policies in reinforcement learning by constraining the magnitude of updates. The algorithm uses a clipped surrogate objective function to balance policy improvement and stability:
\begin{equation}
L^{\text{PPO}} = \mathbb{E}_t \left[\min\left(r_t \cdot A_t, \text{clip}(r_t, 1-\epsilon, 1+\epsilon) \cdot A_t\right)\right],
\end{equation}
where \(r_t = \frac{\pi_{\text{new}}(a_t|s_t)}{\pi_{\text{old}}(a_t|s_t)}\) is the likelihood ratio between the new and old policies and \(A_t\) is the advantage function, computed using Generalized Advantage Estimation (GAE). The clipping mechanism ensures that the updates to the policy remain within a safe range, preventing large deviations from the previous policy and mitigating instability during training.

Additionally, PPO incorporates entropy regularization to encourage exploration and reduce premature convergence. The final objective combines the clipped surrogate loss and entropy regularization, with the latter weighted by a hyperparameter to control its contribution. PPO achieves a balance between exploration and exploitation while avoiding the computational overhead of second-order methods, making it effective in both discrete and continuous action spaces.

VLAC model is a multi-modal large-scale model in which actions are generated through language-based outputs following a predefined template. For instance, an action can be expressed as:

"x: -47mm, y: 19mm, z: 66mm, roll: 14 degrees, pitch: 10 degrees, yaw: 15 degrees, open: 0".

To facilitate autoregressive generation, we define a structured template where non-numeric elements, such as "x:" or "degrees," are directly retrieved from the template. For numeric outputs, the model selects tokens corresponding to numerical values from the token vocabulary. During this process, we record the logits associated with the selected tokens, which represent the probabilities of the corresponding actions. These probabilities are subsequently utilized in the PPO algorithm for policy gradient updates.
The structured template design significantly reduces the number of tokens required for autoregressive generation, thereby improving computational efficiency. By minimizing the token space for generation, this approach not only accelerates inference but also ensures the consistency of the output format, which can be easily parsed into corresponding actions. 

\begin{figure}[t!]
    \centering
    \includegraphics[width=.95\linewidth]{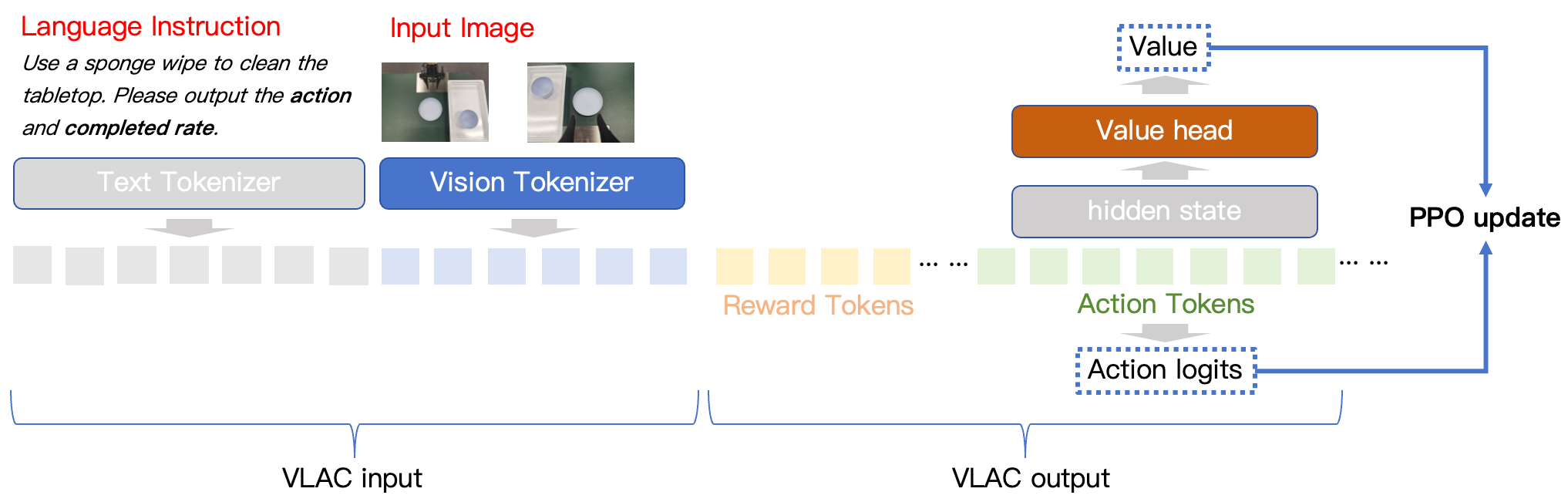}
    \caption{VLAC forward pass generates structured action tokens, reward tokens, and a value head is attached to estimate state value for PPO updates.} %
    \label{fig:vlac_ar}
\end{figure}

Figure~\ref{fig:vlac_ar} summarizes the integration of PPO optimization atop VLAC. After the structured autoregressive action template is decoded, we extract the hidden state (prior to the final token projection) and pass it through a linear value head to obtain $V(s_t)$. Together with the logits from the action tokens, this value estimate is used to compute the PPO loss. The value head is lightly pretrained on a small curated subset of demonstration and early exploration trajectories to reduce initial advantage variance.

\subsubsection{Human-in-the-loop}
\hspace{1.5em}During our experiments, we found that the initial capability of the policy is critically important. In some cases, the policy fails to explore correct actions, and in others, its success rate remains low, requiring targeted practice. Moreover, the training process is often unstable—if the robot fails to explore successfully for a period, the model quickly deteriorates, resulting in a situation where recovery becomes nearly impossible.
To address these issues, we adopted three human-in-the-loop strategies to stabilize the training process:

\textbf{Offline Demonstration Replay}: We implemented an expert demonstration replay buffer, pre-populating it with data collected by human operators. During training, we periodically sample from this dataset and update the model using negative log-likelihood (NLL) loss:  
\begin{equation}
      L_{NLL}(\theta) = -\sum_{(a_t, s_t) \in D_{human}} \log \pi_{\theta}(a_t | s_t)
\end{equation}
We observed that the effectiveness of this approach is highly influenced by the data collection habits of human operators. Before collecting data, the tele-operator observes the policy execution for a period and intuitively supplements the dataset with successful examples for scenarios where the policy has a low success rate.

\textbf{Return and Explore}: During our experiments, we identified scenarios with high failure rates, such as specific initial positions in tasks like picking up a bowl or sweeping trash. To address these cases, we introduced targeted training: while the robot operates, the tele-operator observes positions where the policy frequently fails and manually resets the robot and objects to these states to begin exploration.

\textbf{Human Guided Explore}: Even with reset points, the policy often fails to explore the correct behavioral trajectories required to complete the task. For such cases, we ask the tele-operator to provide demonstration data, which is then added to the aforementioned human replay buffer to facilitate efficient exploration. This approach significantly accelerates the model's ability to learn key behaviors needed for task completion.

\section{Experiments}
\label{sec:exp}
\subsection{Implementation Details}

\hspace{1.5em}During the pre-training phase of the VLAC model, we used a batch size of 3200 and set the maximum learning rate to 8e-4. 
In the real-world RL experiments, we adopt a 2B VLAC model as the actor, i.e., policy model, and use a 8B VLAC model as the "critic" as introduced above. The "critic" plays two roles: computing the reward for each step and indicating whether the task is done. It is worth noted that the normal critic which used to compute GAE is simplely a deep neural network. The capability of prior policy model is critical in real-world RL experiments, thus we collected nearly 100 trajectories by tele-operation and pretrain the policy model as base model. Since the inference and training are asynchronous, the actor occupies two GPU to speed up the training procedure. The "critic" uses a single GPU for computing reward and done signal. For all the tasks, the observation settings are the same. The observation contains instruction, one front camera image, robot endpose position. The robot in our real-world experiment is AGILE PiPER and it is controlled via a 7-DOF end effector based on the delta pose mechanism.
We use the PPO algorithm to train the RL policy and the baseline in experimental results is just implementing the PPO algorithm. Other methods are based on PPO and introduce some new mechanisms, such as \textbf{Return then explore}, \textbf{Human guided explore}, etc.

\subsection{Evaluation Datasets}
\hspace{1.5em}To evaluate our VLAC model to understand task progress, especially its generalization to out-of-distribution scenarios such as new scenes, new tasks, and new entities, we conducted tests not only on the test sets included in the training datasets, but also on six additional datasets that were not seen during training. These include both human operation datasets and datasets containing failure processes. Specifically, in addition to the Bridge~\cite{walke2023bridgedata} and Droid~\cite{khazatsky2024droid} datasets from the training set, we selected RT1~\cite{brohan2022rt}, RoboNet~\cite{dasari2019robonet}, Dobb-E~\cite{shafiullah2023bringing}, RH20T~\cite{fang2023rh20t}, EgoDex~\cite{hoque2025egodex}, and RoboFAC~\cite{lu2025robofac} for evaluation.

Dobb-E features a unique gripper and only provides gripper-perspective views, making it suitable for testing cross-entity and cross-viewpoint generalization.
RoboNet lacks language annotations and does not exhibit smooth temporal structure, so it should show poor task and temporal correlation in the absence of reference examples.
EgoDex is a human hand manipulation dataset, which can be used to evaluate general task process understanding and the model’s compatibility with dexterous hand tasks.
RoboFAC contains two subsets: one with successful task processes and one with failed processes, providing a direct way to evaluate the model’s progress understanding ability.

\subsection{Evaluation Metrics}
\hspace{1.5em}Following the GVL~\cite{ma2024vision}, we use Value-Order Correlation (VOC) as an evaluation metric for task progress understanding. This metric computes the rank correlation between the predicted values and the chronological order of the input expert video:
\begin{equation}
\begin{cases}
\mathrm{VOC} = \text{rank-correlation}(\mathrm{argsort}(v_1, \ldots, v_T); \mathrm{arange}(T))\\
v_i=v_{i-\Delta t}+c_{i-\Delta t,i}(1-v_{i-\Delta t})\\
v_0=0
\end{cases}
\end{equation}
VOC ranges from -1 to 1. Higher VOC scores indicate better task completion, with task progression increasing over time. A good critic model should achieve high VOC scores when evaluated on expert videos.
To better evaluate pairwise methods, we additionally construct Value-Reversed-Order Correlation (VROC). During testing, the entire sequence is reversed, so the order of pairwise frames is inverted and the predicted values should also be reversed. The closer the VROC score is to the VOC score, the better and more stable the model’s performance.
We further define VOC-F1 as:
\begin{equation}
\mathrm{VOC\text{-}F1} = 2 \cdot \frac{\mathrm{VOC} \cdot \mathrm{VROC}}{\mathrm{VOC} + \mathrm{VROC}}
\end{equation}
which comprehensively evaluates the correlation of task progression and temporal order in the video.
Additionally, to evaluate the fine-grained performance of the critic model, we use Negative Rate (NR), which measures the proportion of reversed process pairs:
\begin{equation}
\mathrm{NR} = \frac{N(c_{i,i+\Delta t}<0)}{N}
\end{equation}
where $N(c_{i,i+\Delta t}<0)$ is the number of negative evaluations, and $N$ is the total number of evaluations. This metric reflects how many actions in the video do not contribute to task progression and is also an important indicator of the quality of a trajectory.

Following the $\pi_{0.5}$~\cite{intelligence2504pi0}, we success rate and task progress as evaluation metrics for action generation. Task progress is evaluated by humans, allowing for more detailed progress assessments when the task is not fully completed, and providing a finer evaluation of the model's manipulation ability.

For real-world RL experiments, we mainly investigate the success rate and learning efficiency. The success rate reflects the capability of the policy and it is evaluated by running 10 trials. The learning efficiency can be observed from the success rate curve during the training procedure.

\subsection{VLAC Critic Performance}

\begin{figure}[htbp]
    \centering
    \includegraphics[width=.95\linewidth]{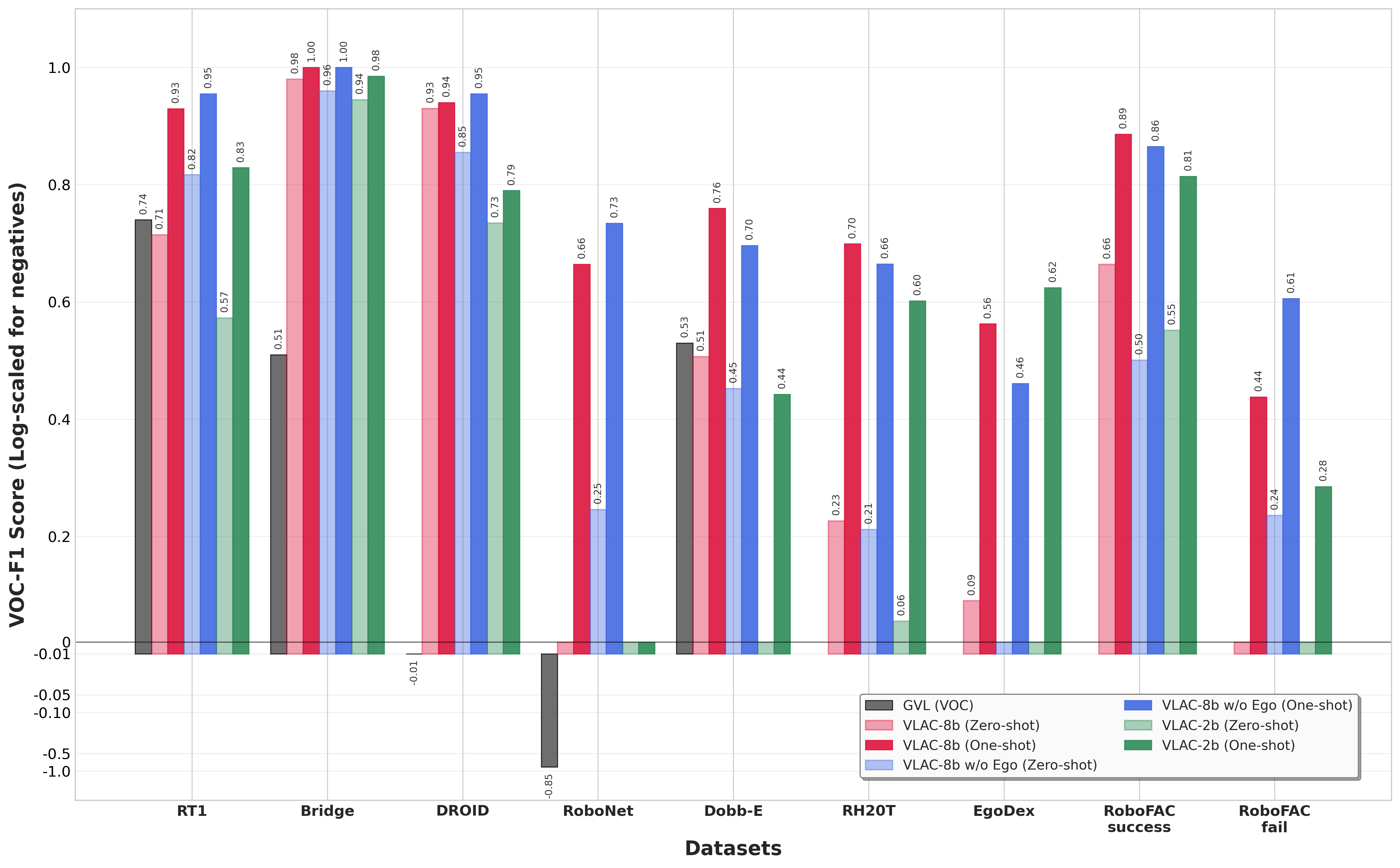}
    \caption{VOC-F1 Performance Comparison Across Different Models and Datasets.} %
    \label{fig:VLAC-VOC}
\end{figure} 


The VLAC model trained on public robotic manipulation data mainly includes bridge, droid, roboset, fmb, AgiBot World, excluding the RT1 and other datasets, and the robotic arm entities, scenes, and tasks in these datasets are almost different from the above datasets. We will conduct further large-scale validation on these datasets. 
We trained three models: the VLAC-8b model with 8 billion parameters, the VLAC-8b w/o Ego model (which was trained without the Ego4D dataset), and the VLAC-2b model with 2 billion parameters. The overall performance across the eight datasets is shown in Figure~\ref{fig:VLAC-VOC}. Our model demonstrates strong results on in-distribution datasets (Bridge and Droid), as well as robust generalization on out-of-distribution datasets. Notably, under one-shot conditions, the model’s powerful contextual reasoning significantly improves its ability to assess task progression. For example, in the RT1 dataset, where the tasks, scenes, and robotic arms differ greatly from those in the training data, the VOC-F1 still reaches 0.95, indicating highly accurate task process prediction. On the RoboNet dataset, which lacks language annotations and does not exhibit smooth temporal structure, the VOC-F1 is 0 in the zero-shot setting—reflecting the model's correct reliance on language descriptions for task progression. However, when provided with examples, the one-shot performance improves dramatically, further highlighting the model’s strong contextual learning capabilities. For the EgoDex dataset, which consists of human demonstration data, we observe that even without incorporating the Ego4D dataset, the model can still leverage context to understand task processes. After including Ego4D in training, this capability is significantly enhanced. The RoboFAC dataset contains both successful and failed task executions. Our method clearly distinguishes between these two types of trajectories, achieving a VOC-F1 of 0.89 on successful videos and only 0.44 on failed ones. This demonstrates VLAC’s strong ability to identify erroneous actions. Furthermore, for RoboFAC, the model trained with Ego4D data shows an even greater gap between successful and failed videos, indicating that human video data provides significant benefits for embodied task process understanding. The specific experimental results are shown in the Table~\ref{tab:critic_table}. 



\begin{table}[t]
\centering
\renewcommand{\arraystretch}{2}
\adjustbox{width=\textwidth,center}
{
\begin{tabular}{|l|l|llllll|llllll|llllll|}
\hline
\rowcolor[HTML]{D8E6FA} 
\multicolumn{1}{|c|}{\cellcolor[HTML]{EFEFEF}}                          & \multicolumn{1}{c|}{\cellcolor[HTML]{D8E6FA}\begin{tabular}[c]{@{}c@{}}GVL\\ (Gemini-1.5-Pro)\end{tabular}} & \multicolumn{6}{c|}{\cellcolor[HTML]{D8E6FA}VLAC-8b w/o Ego (InternVL-8b)}                                                                                                                                                                                                                                  & \multicolumn{6}{c|}{\cellcolor[HTML]{D8E6FA}VLAC-8b (InternVL-8b)}                                                                                                                                                                                                                                                    & \multicolumn{6}{c|}{\cellcolor[HTML]{D8E6FA}VLAC-2b (InternVL-2b)}                                                                                                                                                                                                                                                                      \\ \hline
\rowcolor[HTML]{EFEFEF} 
\multicolumn{1}{|c|}{\cellcolor[HTML]{EFEFEF}}                          & \multicolumn{1}{c|}{\cellcolor[HTML]{EFEFEF}}                                                               & \multicolumn{3}{c|}{\cellcolor[HTML]{EFEFEF}zero-shot}                                                                                                      & \multicolumn{3}{c|}{\cellcolor[HTML]{EFEFEF}one-shot}                                                                                   & \multicolumn{3}{c|}{\cellcolor[HTML]{EFEFEF}zero-shot}                                                                                                     & \multicolumn{3}{c|}{\cellcolor[HTML]{EFEFEF}one-shot}                                                                                                          & \multicolumn{3}{c|}{\cellcolor[HTML]{EFEFEF}zero-shot}                                                                                                     & \multicolumn{3}{c|}{\cellcolor[HTML]{EFEFEF}one-shot}                                                                                                                            \\ \cline{2-20} 
\multicolumn{1}{|c|}{\multirow{-2}{*}{\cellcolor[HTML]{EFEFEF}Dataset}} & VOC                                                                                                         & \multicolumn{1}{l|}{VOC}                           & \multicolumn{1}{l|}{VROC}                          & \multicolumn{1}{l|}{NR}                           & \multicolumn{1}{l|}{VOC}                                   & \multicolumn{1}{l|}{VROC}                                  & NR            & \multicolumn{1}{l|}{VOC}                          & \multicolumn{1}{l|}{VROC}                          & \multicolumn{1}{l|}{NR}                           & \multicolumn{1}{l|}{VOC}                                   & \multicolumn{1}{l|}{VROC}                                                         & NR            & \multicolumn{1}{l|}{VOC}                           & \multicolumn{1}{l|}{VROC}                         & \multicolumn{1}{l|}{NR}                           & \multicolumn{1}{l|}{VOC}                                & \multicolumn{1}{l|}{VROC}                                                         & NR                                 \\ \hline
\rowcolor[HTML]{FFFFFF} 
\cellcolor[HTML]{EFEFEF}RT1                                             & 0.74                                                                                                        & \multicolumn{1}{l|}{\cellcolor[HTML]{FFFFFF}0.87}  & \multicolumn{1}{l|}{\cellcolor[HTML]{FFFFFF}0.77}  & \multicolumn{1}{l|}{\cellcolor[HTML]{FFFFFF}0.19} & \multicolumn{1}{l|}{\cellcolor[HTML]{FFFFFF}0.96} & \multicolumn{1}{l|}{\cellcolor[HTML]{FFFFFF}0.95} & 0.11 & \multicolumn{1}{l|}{\cellcolor[HTML]{FFFFFF}0.71} & \multicolumn{1}{l|}{\cellcolor[HTML]{FFFFFF}0.72}  & \multicolumn{1}{l|}{\cellcolor[HTML]{FFFFFF}0.22} & \multicolumn{1}{l|}{\cellcolor[HTML]{FFFFFF}0.91}          & \multicolumn{1}{l|}{\cellcolor[HTML]{FFFFFF}0.95}                        & 0.14          & \multicolumn{1}{l|}{\cellcolor[HTML]{FFFFFF}0.69}  & \multicolumn{1}{l|}{\cellcolor[HTML]{FFFFFF}0.49} & \multicolumn{1}{l|}{\cellcolor[HTML]{FFFFFF}0.22} & \multicolumn{1}{l|}{\cellcolor[HTML]{FFFFFF}0.80}       & \multicolumn{1}{l|}{\cellcolor[HTML]{FFFFFF}0.86}                                 & \cellcolor[HTML]{FFFFFF}0.25       \\ \hline
\rowcolor[HTML]{FFFFFF} 
\cellcolor[HTML]{EFEFEF}Bridge                                          & 0.51                                                                                                        & \multicolumn{1}{l|}{\cellcolor[HTML]{FFFFFF}0.96}  & \multicolumn{1}{l|}{\cellcolor[HTML]{FFFFFF}0.96}  & \multicolumn{1}{l|}{\cellcolor[HTML]{FFFFFF}0.05} & \multicolumn{1}{l|}{\cellcolor[HTML]{FFFFFF}1.00} & \multicolumn{1}{l|}{\cellcolor[HTML]{FFFFFF}1.00} & 0.00 & \multicolumn{1}{l|}{\cellcolor[HTML]{FFFFFF}0.97} & \multicolumn{1}{l|}{\cellcolor[HTML]{FFFFFF}0.99}  & \multicolumn{1}{l|}{\cellcolor[HTML]{FFFFFF}0.02} & \multicolumn{1}{l|}{\cellcolor[HTML]{FFFFFF}1.00} & \multicolumn{1}{l|}{\cellcolor[HTML]{FFFFFF}1.00}                        & 0.00 & \multicolumn{1}{l|}{\cellcolor[HTML]{FFFFFF}0.94}  & \multicolumn{1}{l|}{\cellcolor[HTML]{FFFFFF}0.95} & \multicolumn{1}{l|}{\cellcolor[HTML]{FFFFFF}0.08} & \multicolumn{1}{l|}{\cellcolor[HTML]{FFFFFF}0.98}       & \multicolumn{1}{l|}{\cellcolor[HTML]{FFFFFF}0.99}                                 & \cellcolor[HTML]{FFFFFF}0.05       \\ \hline
\rowcolor[HTML]{FFFFFF} 
\cellcolor[HTML]{EFEFEF}DROID                                           & -0.01                                                                                                       & \multicolumn{1}{l|}{\cellcolor[HTML]{FFFFFF}0.85}  & \multicolumn{1}{l|}{\cellcolor[HTML]{FFFFFF}0.86}  & \multicolumn{1}{l|}{\cellcolor[HTML]{FFFFFF}0.09} & \multicolumn{1}{l|}{\cellcolor[HTML]{FFFFFF}0.96} & \multicolumn{1}{l|}{\cellcolor[HTML]{FFFFFF}0.95} & 0.06 & \multicolumn{1}{l|}{\cellcolor[HTML]{FFFFFF}0.92} & \multicolumn{1}{l|}{\cellcolor[HTML]{FFFFFF}0.94}  & \multicolumn{1}{l|}{\cellcolor[HTML]{FFFFFF}0.04} & \multicolumn{1}{l|}{\cellcolor[HTML]{FFFFFF}0.93}          & \multicolumn{1}{l|}{\cellcolor[HTML]{FFFFFF}{\color[HTML]{333333} 0.95}} & 0.06          & \multicolumn{1}{l|}{\cellcolor[HTML]{FFFFFF}0.75}  & \multicolumn{1}{l|}{\cellcolor[HTML]{FFFFFF}0.72} & \multicolumn{1}{l|}{\cellcolor[HTML]{FFFFFF}0.13} & \multicolumn{1}{l|}{\cellcolor[HTML]{FFFFFF}0.79}       & \multicolumn{1}{l|}{\cellcolor[HTML]{FFFFFF}{\color[HTML]{333333} 0.79}}          & \cellcolor[HTML]{FFFFFF}0.11       \\ \hline
\rowcolor[HTML]{FFFFFF} 
\cellcolor[HTML]{EFEFEF}RoboNet                                         & -0.85                                                                                                       & \multicolumn{1}{l|}{\cellcolor[HTML]{FFFFFF}0.20}  & \multicolumn{1}{l|}{\cellcolor[HTML]{FFFFFF}0.32}  & \multicolumn{1}{l|}{\cellcolor[HTML]{FFFFFF}0.50} & \multicolumn{1}{l|}{\cellcolor[HTML]{FFFFFF}0.76} & \multicolumn{1}{l|}{\cellcolor[HTML]{FFFFFF}0.71}          & 0.23 & \multicolumn{1}{l|}{\cellcolor[HTML]{FFFFFF}NAN}  & \multicolumn{1}{l|}{\cellcolor[HTML]{FFFFFF}NAN}   & \multicolumn{1}{l|}{\cellcolor[HTML]{FFFFFF}NAN}  & \multicolumn{1}{l|}{\cellcolor[HTML]{FFFFFF}0.59}          & \multicolumn{1}{l|}{\cellcolor[HTML]{FFFFFF}{\color[HTML]{333333} 0.76}} & 0.31          & \multicolumn{1}{l|}{\cellcolor[HTML]{FFFFFF}NAN}   & \multicolumn{1}{l|}{\cellcolor[HTML]{FFFFFF}NAN}  & \multicolumn{1}{l|}{\cellcolor[HTML]{FFFFFF}NAN}  & \multicolumn{1}{l|}{\cellcolor[HTML]{FFFFFF}-0.04}      & \multicolumn{1}{l|}{\cellcolor[HTML]{FFFFFF}{\color[HTML]{333333} 0.81}} & \cellcolor[HTML]{FFFFFF}0.52       \\ \hline
\rowcolor[HTML]{FFFFFF} 
\cellcolor[HTML]{EFEFEF}Dobb-E                                          & 0.53                                                                                                        & \multicolumn{1}{l|}{\cellcolor[HTML]{FFFFFF}0.49}  & \multicolumn{1}{l|}{\cellcolor[HTML]{FFFFFF}0.42}  & \multicolumn{1}{l|}{\cellcolor[HTML]{FFFFFF}0.39} & \multicolumn{1}{l|}{\cellcolor[HTML]{FFFFFF}0.75} & \multicolumn{1}{l|}{\cellcolor[HTML]{FFFFFF}0.65}          & 0.28          & \multicolumn{1}{l|}{\cellcolor[HTML]{FFFFFF}0.47} & \multicolumn{1}{l|}{\cellcolor[HTML]{FFFFFF}0.55}  & \multicolumn{1}{l|}{\cellcolor[HTML]{FFFFFF}0.34} & \multicolumn{1}{l|}{\cellcolor[HTML]{FFFFFF}0.74}          & \multicolumn{1}{l|}{\cellcolor[HTML]{FFFFFF}0.78}                        & 0.26 & \multicolumn{1}{l|}{\cellcolor[HTML]{FFFFFF}-0.04} & \multicolumn{1}{l|}{\cellcolor[HTML]{FFFFFF}0.43} & \multicolumn{1}{l|}{\cellcolor[HTML]{FFFFFF}0.36} & \multicolumn{1}{l|}{\cellcolor[HTML]{FFFFFF}0.37}       & \multicolumn{1}{l|}{\cellcolor[HTML]{FFFFFF}0.55}                                 & \cellcolor[HTML]{FFFFFF}0.31       \\ \hline
\rowcolor[HTML]{FFFFFF} 
\cellcolor[HTML]{EFEFEF}RH20T                                           & none                                                                                                        & \multicolumn{1}{l|}{\cellcolor[HTML]{FFFFFF}0.24}  & \multicolumn{1}{l|}{\cellcolor[HTML]{FFFFFF}0.19}  & \multicolumn{1}{l|}{\cellcolor[HTML]{FFFFFF}0.34} & \multicolumn{1}{l|}{\cellcolor[HTML]{FFFFFF}0.68} & \multicolumn{1}{l|}{\cellcolor[HTML]{FFFFFF}0.65}          & 0.21 & \multicolumn{1}{l|}{\cellcolor[HTML]{FFFFFF}0.17} & \multicolumn{1}{l|}{\cellcolor[HTML]{FFFFFF}0.34}  & \multicolumn{1}{l|}{\cellcolor[HTML]{FFFFFF}0.32} & \multicolumn{1}{l|}{\cellcolor[HTML]{FFFFFF}0.64}          & \multicolumn{1}{l|}{\cellcolor[HTML]{FFFFFF}0.77}                                 & 0.22          & \multicolumn{1}{l|}{\cellcolor[HTML]{FFFFFF}0.03}  & \multicolumn{1}{l|}{\cellcolor[HTML]{FFFFFF}0.40} & \multicolumn{1}{l|}{\cellcolor[HTML]{FFFFFF}0.32} & \multicolumn{1}{l|}{\cellcolor[HTML]{FFFFFF}0.49}       & \multicolumn{1}{l|}{\cellcolor[HTML]{FFFFFF}0.78}                        & \cellcolor[HTML]{FFFFFF}0.25       \\ \hline
\rowcolor[HTML]{FFFFFF} 
\cellcolor[HTML]{EFEFEF}EgoDex                                          & none                                                                                                        & \multicolumn{1}{l|}{\cellcolor[HTML]{FFFFFF}0.58}  & \multicolumn{1}{l|}{\cellcolor[HTML]{FFFFFF}-0.41} & \multicolumn{1}{l|}{\cellcolor[HTML]{FFFFFF}0.29} & \multicolumn{1}{l|}{\cellcolor[HTML]{FFFFFF}0.64} & \multicolumn{1}{l|}{\cellcolor[HTML]{FFFFFF}0.36}          & 0.32 & \multicolumn{1}{l|}{\cellcolor[HTML]{FFFFFF}0.05} & \multicolumn{1}{l|}{\cellcolor[HTML]{FFFFFF}0.48}  & \multicolumn{1}{l|}{\cellcolor[HTML]{FFFFFF}0.48} & \multicolumn{1}{l|}{\cellcolor[HTML]{FFFFFF}0.44}          & \multicolumn{1}{l|}{\cellcolor[HTML]{FFFFFF}0.78}                       & 0.38          & \multicolumn{1}{l|}{\cellcolor[HTML]{FFFFFF}0.00}  & \multicolumn{1}{l|}{\cellcolor[HTML]{FFFFFF}0.59} & \multicolumn{1}{l|}{\cellcolor[HTML]{FFFFFF}0.47} & \multicolumn{1}{l|}{\cellcolor[HTML]{FFFFFF}0.57}       & \multicolumn{1}{l|}{\cellcolor[HTML]{FFFFFF}0.69}                                 & \cellcolor[HTML]{FFFFFF}0.36       \\ \hline
\rowcolor[HTML]{FFFFFF} 
\cellcolor[HTML]{EFEFEF}RoboFAC-success                                 & none                                                                                                        & \multicolumn{1}{l|}{\cellcolor[HTML]{FFFFFF}0.700} & \multicolumn{1}{l|}{\cellcolor[HTML]{FFFFFF}0.39}  & \multicolumn{1}{l|}{\cellcolor[HTML]{FFFFFF}0.26} & \multicolumn{1}{l|}{\cellcolor[HTML]{FFFFFF}0.86} & \multicolumn{1}{l|}{\cellcolor[HTML]{FFFFFF}0.87}          & 0.21          & \multicolumn{1}{l|}{\cellcolor[HTML]{FFFFFF}0.76} & \multicolumn{1}{l|}{\cellcolor[HTML]{FFFFFF}0.59}  & \multicolumn{1}{l|}{\cellcolor[HTML]{FFFFFF}0.14} & \multicolumn{1}{l|}{\cellcolor[HTML]{FFFFFF}0.83}          & \multicolumn{1}{l|}{\cellcolor[HTML]{FFFFFF}0.95}                        & 0.20 & \multicolumn{1}{l|}{\cellcolor[HTML]{FFFFFF}0.46}  & \multicolumn{1}{l|}{\cellcolor[HTML]{FFFFFF}0.69} & \multicolumn{1}{l|}{\cellcolor[HTML]{FFFFFF}0.20} & \multicolumn{1}{l|}{\cellcolor[HTML]{FFFFFF}0.73}       & \multicolumn{1}{l|}{\cellcolor[HTML]{FFFFFF}0.92}                                 & \cellcolor[HTML]{FFFFFF}0.29       \\ \hline
\rowcolor[HTML]{FFFFFF} 
\cellcolor[HTML]{EFEFEF}RoboFAC-fail                                    & none                                                                                                        & \multicolumn{1}{l|}{\cellcolor[HTML]{FFFFFF}0.45}  & \multicolumn{1}{l|}{\cellcolor[HTML]{FFFFFF}0.16}  & \multicolumn{1}{l|}{\cellcolor[HTML]{FFFFFF}0.34} & \multicolumn{1}{l|}{\cellcolor[HTML]{FFFFFF}0.66}          & \multicolumn{1}{l|}{\cellcolor[HTML]{FFFFFF}0.56}          & 0.32          & \multicolumn{1}{l|}{\cellcolor[HTML]{FFFFFF}0.30} & \multicolumn{1}{l|}{\cellcolor[HTML]{FFFFFF}-0.06} & \multicolumn{1}{l|}{\cellcolor[HTML]{FFFFFF}0.22} & \multicolumn{1}{l|}{\cellcolor[HTML]{FFFFFF}0.41}          & \multicolumn{1}{l|}{\cellcolor[HTML]{FFFFFF}{0.47}}                           & 0.35          & \multicolumn{1}{l|}{\cellcolor[HTML]{FFFFFF}-0.05} & \multicolumn{1}{l|}{\cellcolor[HTML]{FFFFFF}0.34} & \multicolumn{1}{l|}{\cellcolor[HTML]{FFFFFF}0.26} & \multicolumn{1}{l|}{\cellcolor[HTML]{FFFFFF}{0.19}} & \multicolumn{1}{l|}{\cellcolor[HTML]{FFFFFF}0.57}                                 & \cellcolor[HTML]{FFFFFF}{0.43} \\ \hline
\end{tabular}
}
\caption{Performance of VLAC's task progressing understanding across entities, scenarios, and tasks.}
\label{tab:critic_table}
\end{table}

\begin{figure}[t]
    \centering
    \includegraphics[width=1.0\linewidth]{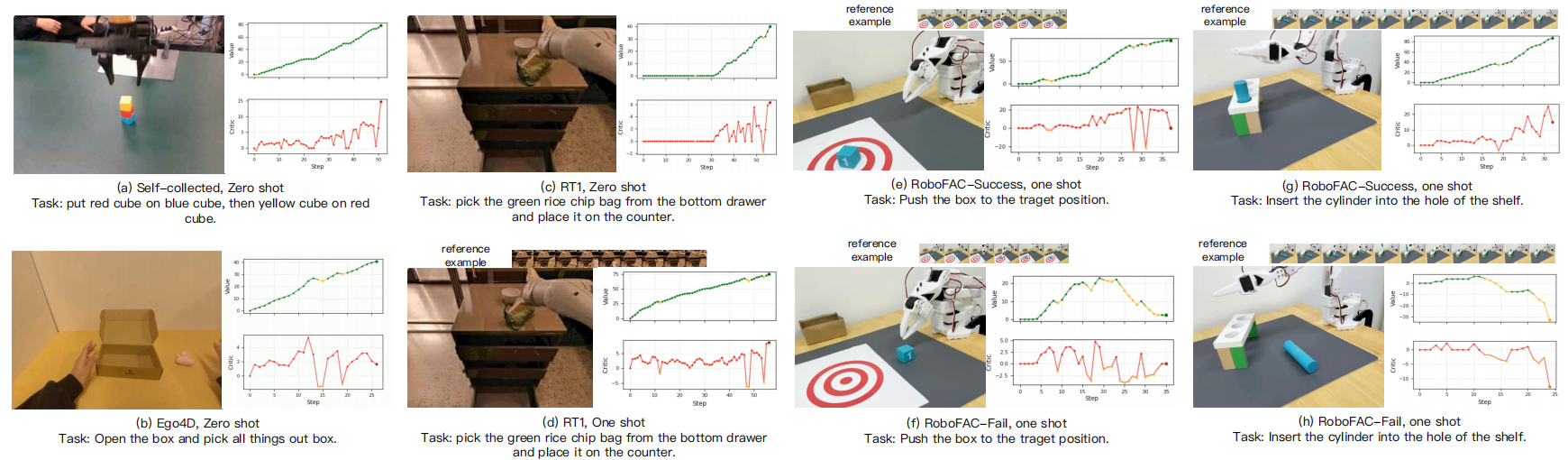}
    \caption{Example results of VLAC for task progress understanding.} %
    \label{fig:case study}
    \vspace{-2mm}
\end{figure} 

We show some example results in Figure~\ref{fig:case study}, where (b–h) illustrate performance on datasets cross different entities, scenes, and tasks, demonstrating the strong generalization capabilities of the proposed model. More specifically, Figure~\ref{fig:case study} (b) highlights the model’s understanding of the human dataset. Figures~\ref{fig:case study} (c) and (d) show zero-shot and one-shot results for the same process, illustrating how in-context learning enhances the model’s ability to comprehend new task processes. Figures~\ref{fig:case study} (e) and (g) depict successful task processes, while Figures~\ref{fig:case study} (f) and (h) present the corresponding failed task processes, indicating that the model can clearly distinguish between successful and unsuccessful processes. These examples show our model can robustly generalize across heterogeneous embodiments, scenes, and tasks, leveraging minimal in-context signals to accurately infer, monitor, and contrast fine-grained task progress, and even differentiate subtle failure modes from successful executions.

\subsection{VLAC Actor Performance}

\begin{table}[htbp]
\centering
\renewcommand{\arraystretch}{2}
\adjustbox{width=\textwidth,center}
{
\begin{tabular}{|l|ll|ll|ll|ll|ll|ll|}
\hline
\rowcolor[HTML]{D8E6FA} 
\multicolumn{1}{|c|}{\cellcolor[HTML]{EFEFEF}}                          & \multicolumn{2}{c|}{\cellcolor[HTML]{D8E6FA}Open Cooker}                  & \multicolumn{2}{c|}{\cellcolor[HTML]{D8E6FA}Pick\&Place Bowl}            & \multicolumn{2}{c|}{\cellcolor[HTML]{D8E6FA}Unfold Mat}                 & \multicolumn{2}{c|}{\cellcolor[HTML]{D8E6FA}Desktop Sweep}                  & \multicolumn{2}{c|}{\cellcolor[HTML]{D8E6FA}Rice Transfer}         & \multicolumn{2}{c|}{\cellcolor[HTML]{D8E6FA}\textbf{Average}}                \\ \hline
\rowcolor[HTML]{EFEFEF} 
\multicolumn{1}{|c|}{\cellcolor[HTML]{EFEFEF}Method} & \multicolumn{1}{l|}{\cellcolor[HTML]{EFEFEF}\makecell{task\\progress}} & \cellcolor[HTML]{EFEFEF}\makecell{success\\rate} & \multicolumn{1}{l|}{\cellcolor[HTML]{EFEFEF}\makecell{task\\progress}} & \cellcolor[HTML]{EFEFEF}\makecell{success\\rate} & \multicolumn{1}{l|}{\cellcolor[HTML]{EFEFEF}\makecell{task\\progress}} & \cellcolor[HTML]{EFEFEF}\makecell{success\\rate} & \multicolumn{1}{l|}{\cellcolor[HTML]{EFEFEF}\makecell{task\\progress}} & \cellcolor[HTML]{EFEFEF}\makecell{success\\rate} & \multicolumn{1}{l|}{\cellcolor[HTML]{EFEFEF}\makecell{task\\progress}} & \cellcolor[HTML]{EFEFEF}\makecell{success\\rate} & \multicolumn{1}{l|}{\cellcolor[HTML]{EFEFEF}\makecell{task\\progress}}   & \cellcolor[HTML]{EFEFEF}\makecell{success\\rate}  \\ \hline
\rowcolor[HTML]{FFFFFF} 
\cellcolor[HTML]{EFEFEF}Pi0                    & \multicolumn{1}{l|}{\cellcolor[HTML]{FFFFFF}85\%}          & \cellcolor[HTML]{FFFFFF}85\%         & \multicolumn{1}{l|}{\cellcolor[HTML]{FFFFFF}54\%}          & \cellcolor[HTML]{FFFFFF}40\%         & \multicolumn{1}{l|}{\cellcolor[HTML]{FFFFFF}32\%}          & \cellcolor[HTML]{FFFFFF}0\%          & \multicolumn{1}{l|}{\cellcolor[HTML]{FFFFFF}30\%}          & \cellcolor[HTML]{FFFFFF}10\%         & \multicolumn{1}{l|}{\cellcolor[HTML]{FFFFFF}45\%}          & \cellcolor[HTML]{FFFFFF}30\%         & \multicolumn{1}{l|}{\cellcolor[HTML]{FFFFFF}49.2\%}          & \cellcolor[HTML]{FFFFFF}27\%          \\ \hline
\rowcolor[HTML]{FFFFFF} 
\cellcolor[HTML]{EFEFEF}Pi0+Lighting Transfer  & \multicolumn{1}{l|}{\cellcolor[HTML]{FFFFFF}0\%}           & \cellcolor[HTML]{FFFFFF}0\%          & \multicolumn{1}{l|}{\cellcolor[HTML]{FFFFFF}7.5\%}         & \cellcolor[HTML]{FFFFFF}0\%          & \multicolumn{1}{l|}{\cellcolor[HTML]{FFFFFF}35.5\%}        & \cellcolor[HTML]{FFFFFF}5\%          & \multicolumn{1}{l|}{\cellcolor[HTML]{FFFFFF}31.5\%}        & \cellcolor[HTML]{FFFFFF}10\%         & \multicolumn{1}{l|}{\cellcolor[HTML]{FFFFFF}9\%}           & \cellcolor[HTML]{FFFFFF}0\%          & \multicolumn{1}{l|}{\cellcolor[HTML]{FFFFFF}16.7\%}          & \cellcolor[HTML]{FFFFFF}3\%           \\ \hline
\rowcolor[HTML]{FFFFFF} 
\cellcolor[HTML]{EFEFEF}VLAC w/o pretrain      & \multicolumn{1}{l|}{\cellcolor[HTML]{FFFFFF}50\%}          & \cellcolor[HTML]{FFFFFF}50\%         & \multicolumn{1}{l|}{\cellcolor[HTML]{FFFFFF}0\%}           & \cellcolor[HTML]{FFFFFF}0\%          & \multicolumn{1}{l|}{\cellcolor[HTML]{FFFFFF}2.5\%}         & \cellcolor[HTML]{FFFFFF}0\%          & \multicolumn{1}{l|}{\cellcolor[HTML]{FFFFFF}0\%}           & \cellcolor[HTML]{FFFFFF}0\%          & \multicolumn{1}{l|}{\cellcolor[HTML]{FFFFFF}56\%}          & \cellcolor[HTML]{FFFFFF}30\%         & \multicolumn{1}{l|}{\cellcolor[HTML]{FFFFFF}21.7\%}          & \cellcolor[HTML]{FFFFFF}16\%          \\ \hline
\rowcolor[HTML]{FFFFFF} 
\cellcolor[HTML]{EFEFEF}VLAC                   & \multicolumn{1}{l|}{\cellcolor[HTML]{FFFFFF}90\%}          & \cellcolor[HTML]{FFFFFF}90\%         & \multicolumn{1}{l|}{\cellcolor[HTML]{FFFFFF}85\%}          & \cellcolor[HTML]{FFFFFF}85\%         & \multicolumn{1}{l|}{\cellcolor[HTML]{FFFFFF}91\%}          & \cellcolor[HTML]{FFFFFF}85\%         & \multicolumn{1}{l|}{\cellcolor[HTML]{FFFFFF}62.5\%}        & \cellcolor[HTML]{FFFFFF}40\%         & \multicolumn{1}{l|}{\cellcolor[HTML]{FFFFFF}84.5\%}        & \cellcolor[HTML]{FFFFFF}75\%         & \multicolumn{1}{l|}{\cellcolor[HTML]{FFFFFF}\textbf{82.5\%}} & \cellcolor[HTML]{FFFFFF}\textbf{75\%} \\ \hline
\rowcolor[HTML]{FFFFFF} 
\cellcolor[HTML]{EFEFEF}VLAC+Lighting Transfer & \multicolumn{1}{l|}{\cellcolor[HTML]{FFFFFF}85\%}          & \cellcolor[HTML]{FFFFFF}85\%         & \multicolumn{1}{l|}{\cellcolor[HTML]{FFFFFF}65\%}          & \cellcolor[HTML]{FFFFFF}65\%         & \multicolumn{1}{l|}{\cellcolor[HTML]{FFFFFF}72.5\%}        & \cellcolor[HTML]{FFFFFF}60\%         & \multicolumn{1}{l|}{\cellcolor[HTML]{FFFFFF}49.5\%}        & \cellcolor[HTML]{FFFFFF}25\%         & \multicolumn{1}{l|}{\cellcolor[HTML]{FFFFFF}75.5\%}        & \cellcolor[HTML]{FFFFFF}50\%         & \multicolumn{1}{l|}{\cellcolor[HTML]{FFFFFF}69.5\%}          & \cellcolor[HTML]{FFFFFF}57\%          \\ \hline
\rowcolor[HTML]{FFFFFF} 
\cellcolor[HTML]{EFEFEF}VLAC+Scene Transfer    & \multicolumn{1}{l|}{\cellcolor[HTML]{FFFFFF}90\%}          & \cellcolor[HTML]{FFFFFF}90\%         & \multicolumn{1}{l|}{\cellcolor[HTML]{FFFFFF}70\%}          & \cellcolor[HTML]{FFFFFF}65\%         & \multicolumn{1}{l|}{\cellcolor[HTML]{FFFFFF}80\%}          & \cellcolor[HTML]{FFFFFF}70\%         & \multicolumn{1}{l|}{\cellcolor[HTML]{FFFFFF}60.0\%}        & \cellcolor[HTML]{FFFFFF}40\%         & \multicolumn{1}{l|}{\cellcolor[HTML]{FFFFFF}68\%}          & \cellcolor[HTML]{FFFFFF}50\%         & \multicolumn{1}{l|}{\cellcolor[HTML]{FFFFFF}73.6\%}          & \cellcolor[HTML]{FFFFFF}63\%          \\ \hline
\end{tabular}
}
\caption{Performance of VLAC's action generation across scenarios.}
\label{tab:actor_table}
\end{table}

To validate the action generation capability of VLAC, we collected 100 samples for each testing task and trained the 8B VLAC model across all tasks. Additionally, since autonomous evolution in the real world often requires exploring different environments independently, the model must possess strong generalization abilities to adapt to scene variations.
To test this, we evaluated the VLAC model's success rate under lighting disturbances and scene changes without requiring extra data collection. As shown in Table~\ref{tab:actor_table}, the "Lighting Transfer" scenario involved turning off fluorescent lights and using colored flashing light sources as disturbances. For the "Scene Transfer" scenario, tests were conducted on two different workbenches located at different sites and different settings, with varying camera perspectives that were not precisely calibrated. The examples of training data and evaluation environments can be seen in Figure~\ref{fig:transfer}.
The results demonstrated the model's robust generalization ability, as it was able to consistently generate reasonable actions even under extreme lighting changes. This makes VLAC a suitable starting point for autonomous evolution in dynamic and unpredictable environments. 
Meanwhile, “VLAC w/o pretrain” refers to the model without task progress pretraining, which leads to a significant drop in success rate. Although the actions generated during testing remain reasonable, the model struggles to accurately determine the current task state. For example, in the pick-and-place task, it may proceed to the placement step even if it has not successfully grasped the object. Our process understanding task enhances the model’s ability to interpret the progression of tasks depicted in images, thereby ensuring a higher success rate at each execution stage.
However, the success rate of tasks varies depending on the difficulty of the tasks, and performance is often partially lost during transfer. To improve the success rate during real-world deployment, further autonomous learning and evolution in actual environments and tasks are necessary.
 
\subsection{Real-world RL Results}
\begin{table}[ht]
\centering
\small
\adjustbox{width=\textwidth,center}
{
\begin{tabular}{l|cccc} 
\toprule[1.5pt]
\multirow{2}{*}{\textbf{Task}} & \multicolumn{4}{c}{\textbf{ Success Rate (\%)}} \\ 
\cline{2-5}  
& Baseline & Return and explore & Human guided explore & Offline Demonstration Replay \\ 
\midrule[0.75pt]
Desktop Sweep Disposal & 90  & 100  & 100 & 100 \\ 
Pick and Place Bowl    & 80  & 90   & 100 & 100 \\ 
Unfold Mat             & 80  & 90   & 90  & 100 \\ 
Rice Scooping and Transfer   & 100 & 80   & 100  & 70  \\ 
\midrule[0.75pt]
Average      & 88 & 95 & 98 &  93 \\ 
\bottomrule[1.5pt]
\end{tabular}
}
\caption{Illustration of the final success rate on the real-world tasks. The success rate is evaluated by running 10 episodes.}
\label{tab:rl_success_rate}
\end{table}

\begin{figure}[h!]
    \centering
    \includegraphics[width=.99\linewidth]{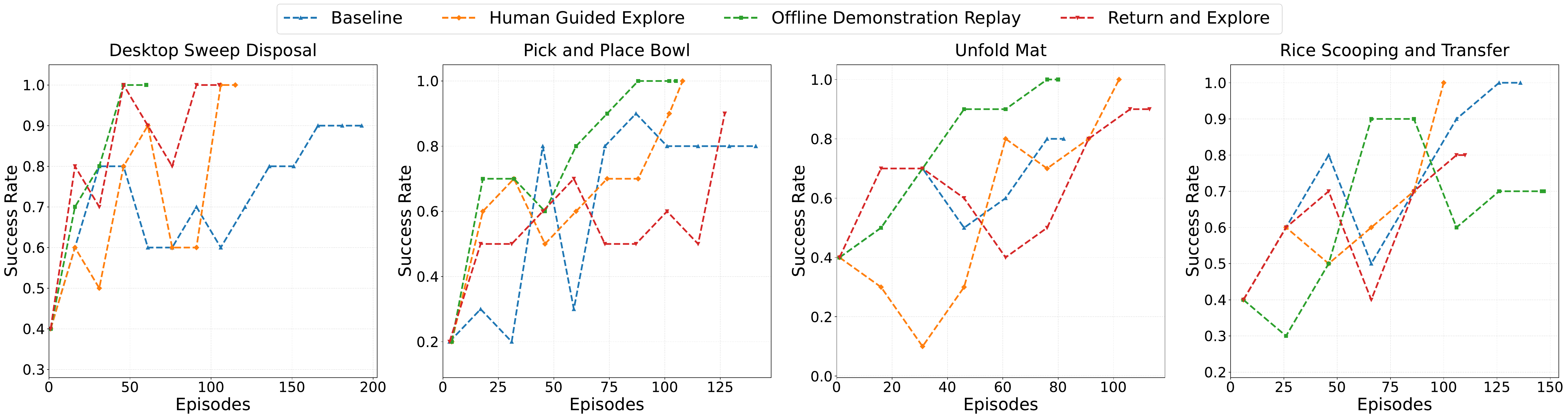}
    \caption{Experimental results of real-world RL on four tasks. This figure demonstrates the success rate for all the methods during the training procedure. Each data point is obtained via running 10 trails.} %
    \label{fig:rl_success_rate}
\end{figure} 
\hspace{1.5em}In this subsection, we mainly investigate how to improve the policy learning performance in different real-world scenarios. Our key insight is to enhance the exploration capabilities of the model. So as to, we introduce \textbf{Offline Demonstration Replay}, \textbf{Return and explore}, \textbf{Human guided explore} in our experiments and the baseline means our real-world rl approach without all these modules.

As shown in  Table~\ref{tab:rl_success_rate}, all these methods can  achieve higher success rate  on all the tasks compared to Baseline. It is worth noted that \textbf{Human guided explore} and \textbf{Offline Demonstration Replay} can  both achieve 100\% success rate four tasks. The experimental results show that human intervention, that is, manually collecting some trajectories that are helpful for task exploration, is important for learning the policy. 
\textbf{Return and explore} only performs slightly better than Baseline and can not achieve similar results compared to above two methods. This may be because choosing  good return points is difficult. Some return points are too easy and the prior policy reaches there  frequently and easily, but some return points are too difficult and the prior policy behaves poorly there and struggles to explore. 

As shown in Figure~\ref{fig:rl_success_rate} , \textbf{Offline Demonstration Replay} learns faster than \textbf{Human guided explore}. This may be because the quality of data we collected in \textbf{Offline Demonstration Replay} is comparable to that of \textbf{Human guided explore}, which enables the model to overcome the difficulties encountered during the exploration process at the beginning of learning. 



\begin{figure}[h!]
    \centering
    \includegraphics[width=.8\linewidth]{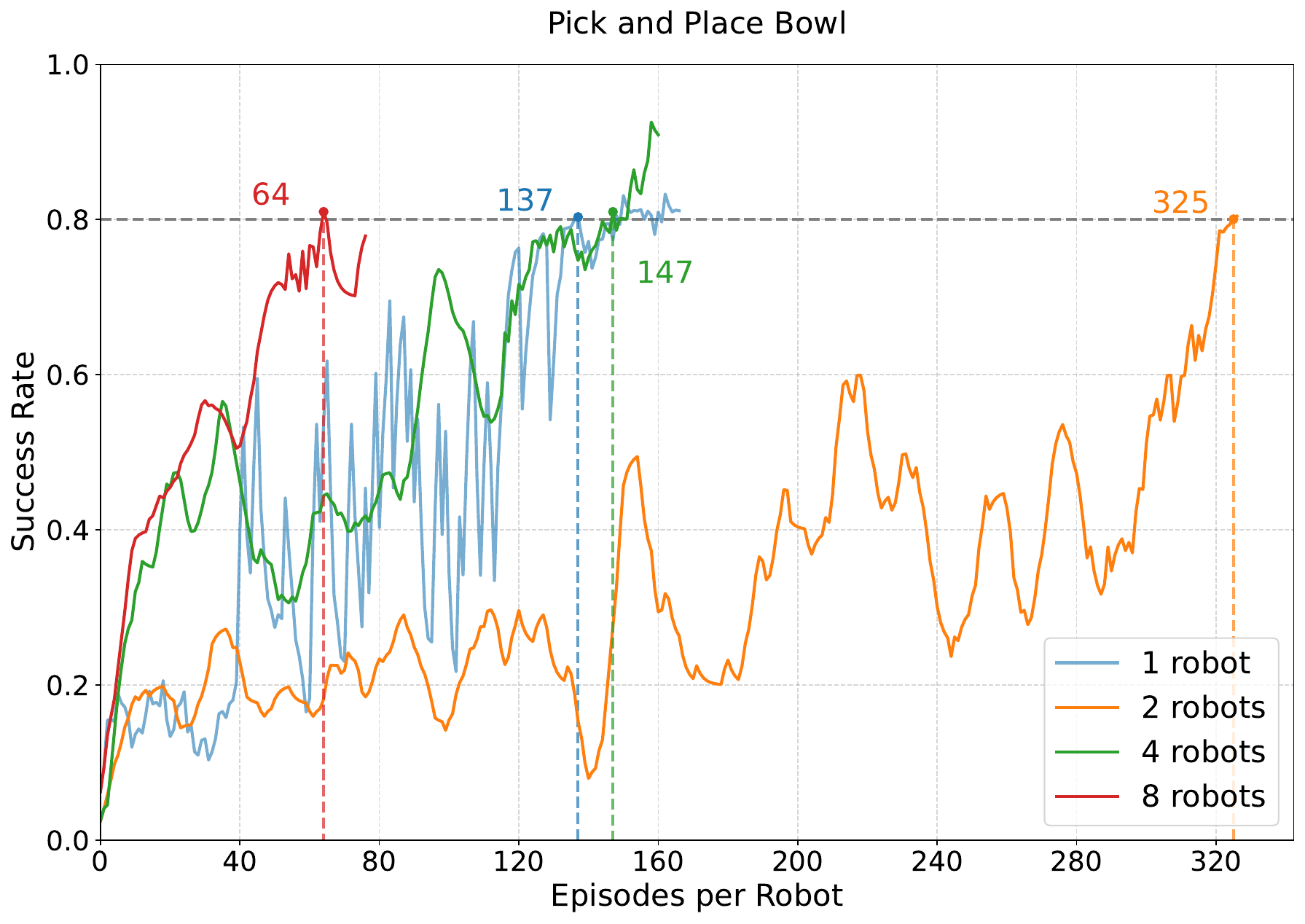}
    \caption{Training performance curves for real-world robots at different scales. The episodes required per robot to reach a high success rate decrease as the number of robots increases from 2 to 4 to 8. } %
    \label{fig:scale_main_result}
\end{figure}

\subsection{ Multi-Robot Scaling}
\hspace{1.5em}We investigate how learning performance scales with the number of robots using a representative tabletop manipulation task (Pick-and-Place Bowl) trained under the Baseline method. We consider robot counts \textit{N}=1,2,4,8 to assess whether a scaling law persists as parallel embodiment increases.

The overall results are shown in Figure ~\ref{fig:scale_main_result}. To quantify learning efficiency, we report the number of episodes per robot required to reach an 80\% success rate (black dots in Figure ~\ref{fig:scale_main_result}. With 8 robots, only 64 episodes per robot are needed, requiring about 24.6 minutes of rollout. Reducing the joint training scale from 8 to 4 robots increases the requirement to 147 episodes per robot (roughly a 2× increase). Further reducing from 8 to 2 robots causes the per‑robot requirement to rise sharply to 325 episodes (over 5× that of 8 robots), taking roughly two hours. Thus, the scaling law appears to hold: increasing the number of robots decreases the data required per robot to reach the same success level. An exception is the single‑robot case, which requires fewer episodes than 2 robots and is comparable to 4 robots. This deviation is mainly due to: (1) its visually static background, which is easier to fit; and (2) heterogeneous backgrounds across multiple robots introducing visual disturbance that slows learning of a more generalized policy.

Unlike in simulation, observations among real-world robots are rarely identical because of hardware differences, background variation, and differing camera viewpoints. Consequently, learning speeds diverge: some robots may still struggle while others have already mastered the skill. To balance overall progress, we apply a dynamic sampling strategy that increases the sampling frequency of data generated by under‑learned robots. Additional details about the environment setup are provided in the appendix.

\section{Discussion and Conclusion}
\label{sec:discussion}
\hspace{1.5em}This work tackles the central challenge of making large Vision-Language-Action (VLA) models actually improve themselves in the physical world. Beyond introducing the VLAC (Vision-Language-Action-Critic) model, the core contribution is a full real-world reinforcement learning (RL) framework that (1) converts general multimodal understanding into dense, signed stepwise intrinsic rewards via calibrated pairwise progress deltas, (2) closes the loop on physical robots with a low‑latency asynchronous infrastructure (actor–critic decoupling, progress-based done detection, reward inference in <0.1s), (3) enables continual self-improvement from both successes and failures, and (4) supports graded, optional human intervention (offline seeding, targeted resets, guided micro-demonstrations) to accelerate exploration. Empirically, across four diverse manipulation tasks the system raises success rates from 30\% to 90\% within 200 real-world interaction episodes, transfers under lighting and scene shifts without extra collection, and reliably separates successful from failing trajectories (high VOC / VROC contrast). The result is a practical recipe showing that large multimodal priors plus structured intrinsic progress feedback can make real-world online RL feasible, data-efficient, and incrementally improvable.

Limitations: (1) Human-in-the-loop dependence: intervention timing, reset state selection, and demonstration curation remain heuristic and operator-specific; no standardized quantitative triggers (e.g., competence plateau detectors, failure mode coverage metrics) are yet defined, limiting reproducibility and automation. (2) RL recipe specificity: the current PPO + autoregressive tokenized delta–EEF action interface is tightly coupled to a discrete semantic action head; it does not directly generalize to diffusion / flow-matching / continuous iterative decoders where reward allocation across denoising steps, progress-to-score alignment, or Q-guided sampling would need new abstractions. (3) Multi-task online instability: simultaneous RL still shows reward scale drift, uneven negative signal density, inter-task gradient interference, and episodic forgetting; no task-adaptive normalization, uncertainty-weighted sampling, gradient conflict mitigation, modular parameter partitioning, or continual distillation safeguards are integrated. Future work will focus on formalizing intervention metrics (coverage, marginal utility curves), designing architecture-agnostic progress/value bridging layers for non-autoregressive action generators, and adding stabilization mechanisms (task-wise reward calibration, uncertainty-driven replay prioritization, gradient surgery, lightweight policy/module distillation) to scale robust multi-task real-world evolution.


\bibliography{refs}

\begin{thebibliography}{65}
\providecommand{\natexlab}[1]{#1}
\providecommand{\url}[1]{\texttt{#1}}
\expandafter\ifx\csname urlstyle\endcsname\relax
  \providecommand{\doi}[1]{doi: #1}\else
  \providecommand{\doi}{doi: \begingroup \urlstyle{rm}\Url}\fi

\bibitem[Bharadhwaj et~al.(2024)Bharadhwaj, Vakil, Sharma, Gupta, Tulsiani, and
  Kumar]{bharadhwaj2024roboagent}
H.~Bharadhwaj, J.~Vakil, M.~Sharma, A.~Gupta, S.~Tulsiani, and V.~Kumar.
\newblock Roboagent: Generalization and efficiency in robot manipulation via
  semantic augmentations and action chunking.
\newblock In \emph{2024 IEEE International Conference on Robotics and
  Automation (ICRA)}, pages 4788--4795. IEEE, 2024.

\bibitem[Biza et~al.(2024)Biza, Weng, Sun, Schmeckpeper, Kelestemur, Ma, Platt,
  van~de Meent, and Wong]{biza2024robot}
O.~Biza, T.~Weng, L.~Sun, K.~Schmeckpeper, T.~Kelestemur, Y.~J. Ma, R.~Platt,
  J.-W. van~de Meent, and L.~L. Wong.
\newblock On-robot reinforcement learning with goal-contrastive rewards.
\newblock \emph{arXiv preprint arXiv:2410.19989}, 2024.

\bibitem[Bjorck et~al.(2025)Bjorck, Casta{\~n}eda, Cherniadev, Da, Ding, Fan,
  Fang, Fox, Hu, Huang, et~al.]{bjorck2025gr00t}
J.~Bjorck, F.~Casta{\~n}eda, N.~Cherniadev, X.~Da, R.~Ding, L.~Fan, Y.~Fang,
  D.~Fox, F.~Hu, S.~Huang, et~al.
\newblock Gr00t n1: An open foundation model for generalist humanoid robots.
\newblock \emph{arXiv preprint arXiv:2503.14734}, 2025.

\bibitem[Black et~al.(2024)Black, Brown, Driess, Esmail, Equi, Finn, Fusai,
  Groom, Hausman, Ichter, et~al.]{black2024pi_0}
K.~Black, N.~Brown, D.~Driess, A.~Esmail, M.~Equi, C.~Finn, N.~Fusai, L.~Groom,
  K.~Hausman, B.~Ichter, et~al.
\newblock $\pi_0 $: A vision-language-action flow model for general robot
  control.
\newblock \emph{arXiv preprint arXiv:2410.24164}, 2024.

\bibitem[Brohan et~al.(2022)Brohan, Brown, Carbajal, Chebotar, Dabis, Finn,
  Gopalakrishnan, Hausman, Herzog, Hsu, et~al.]{brohan2022rt}
A.~Brohan, N.~Brown, J.~Carbajal, Y.~Chebotar, J.~Dabis, C.~Finn,
  K.~Gopalakrishnan, K.~Hausman, A.~Herzog, J.~Hsu, et~al.
\newblock Rt-1: Robotics transformer for real-world control at scale.
\newblock \emph{arXiv preprint arXiv:2212.06817}, 2022.

\bibitem[Brooks et~al.(2023)Brooks, Holynski, and
  Efros]{brooks2023instructpix2pix}
T.~Brooks, A.~Holynski, and A.~A. Efros.
\newblock Instructpix2pix: Learning to follow image editing instructions.
\newblock In \emph{Proceedings of the IEEE/CVF conference on computer vision
  and pattern recognition}, pages 18392--18402, 2023.

\bibitem[Bu et~al.(2025)Bu, Cai, Chen, Cui, Ding, Feng, He, Huang,
  et~al.]{bu2025agibot}
Q.~Bu, J.~Cai, L.~Chen, X.~Cui, Y.~Ding, S.~Feng, X.~He, X.~Huang, et~al.
\newblock Agibot world colosseo: A large-scale manipulation platform for
  scalable and intelligent embodied systems.
\newblock In \emph{2025 IEEE/RSJ International Conference on Intelligent Robots
  and Systems (IROS)}. IEEE, 2025.

\bibitem[Chen et~al.(2024)Chen, Xu, Kirmani, Ichter, Sadigh, Guibas, and
  Xia]{chen2024spatialvlm}
B.~Chen, Z.~Xu, S.~Kirmani, B.~Ichter, D.~Sadigh, L.~Guibas, and F.~Xia.
\newblock Spatialvlm: Endowing vision-language models with spatial reasoning
  capabilities.
\newblock In \emph{Proceedings of the IEEE/CVF Conference on Computer Vision
  and Pattern Recognition}, pages 14455--14465, 2024.

\bibitem[Chen et~al.(2025)Chen, Tian, Liu, Zhou, Li, and Zhao]{chen2025conrft}
Y.~Chen, S.~Tian, S.~Liu, Y.~Zhou, H.~Li, and D.~Zhao.
\newblock Conrft: A reinforced fine-tuning method for vla models via
  consistency policy.
\newblock \emph{arXiv preprint arXiv:2502.05450}, 2025.

\bibitem[Chi et~al.(2023)Chi, Xu, Feng, Cousineau, Du, Burchfiel, Tedrake, and
  Song]{chi2023diffusion}
C.~Chi, Z.~Xu, S.~Feng, E.~Cousineau, Y.~Du, B.~Burchfiel, R.~Tedrake, and
  S.~Song.
\newblock Diffusion policy: Visuomotor policy learning via action diffusion.
\newblock \emph{The International Journal of Robotics Research}, page
  02783649241273668, 2023.

\bibitem[Dasari et~al.(2019)Dasari, Ebert, Tian, Nair, Bucher, Schmeckpeper,
  Singh, Levine, and Finn]{dasari2019robonet}
S.~Dasari, F.~Ebert, S.~Tian, S.~Nair, B.~Bucher, K.~Schmeckpeper, S.~Singh,
  S.~Levine, and C.~Finn.
\newblock Robonet: Large-scale multi-robot learning.
\newblock \emph{arXiv preprint arXiv:1910.11215}, 2019.

\bibitem[Deng et~al.(2025)Deng, Yan, Wei, Ma, Yang, Chen, Zhang, Yang, Zhang,
  Cui, et~al.]{deng2025graspvla}
S.~Deng, M.~Yan, S.~Wei, H.~Ma, Y.~Yang, J.~Chen, Z.~Zhang, T.~Yang, X.~Zhang,
  H.~Cui, et~al.
\newblock Graspvla: a grasping foundation model pre-trained on billion-scale
  synthetic action data.
\newblock \emph{arXiv preprint arXiv:2505.03233}, 2025.

\bibitem[Ding and Jin()]{dingconsistency}
Z.~Ding and C.~Jin.
\newblock Consistency models as a rich and efficient policy class for
  reinforcement learning.
\newblock In \emph{The Twelfth International Conference on Learning
  Representations}.

\bibitem[Fang et~al.(2023)Fang, Fang, Tang, Liu, Wang, Wang, Zhu, and
  Lu]{fang2023rh20t}
H.-S. Fang, H.~Fang, Z.~Tang, J.~Liu, C.~Wang, J.~Wang, H.~Zhu, and C.~Lu.
\newblock Rh20t: A comprehensive robotic dataset for learning diverse skills in
  one-shot.
\newblock \emph{arXiv preprint arXiv:2307.00595}, 2023.

\bibitem[Geng et~al.(2025)Geng, Deng, Bai, Kolter, and He]{geng2025mean}
Z.~Geng, M.~Deng, X.~Bai, J.~Z. Kolter, and K.~He.
\newblock Mean flows for one-step generative modeling.
\newblock \emph{arXiv preprint arXiv:2505.13447}, 2025.

\bibitem[He et~al.(2024)He, Shen, Tan, and Wang]{he2024aligniql}
L.~He, L.~Shen, J.~Tan, and X.~Wang.
\newblock Aligniql: Policy alignment in implicit q-learning through constrained
  optimization.
\newblock \emph{arXiv preprint arXiv:2405.18187}, 2024.

\bibitem[Herzog et~al.(2023)Herzog, Rao, Hausman, Lu, Wohlhart, Yan, Lin,
  Arenas, Xiao, Kappler, et~al.]{herzog2023deep}
A.~Herzog, K.~Rao, K.~Hausman, Y.~Lu, P.~Wohlhart, M.~Yan, J.~Lin, M.~G.
  Arenas, T.~Xiao, D.~Kappler, et~al.
\newblock Deep rl at scale: Sorting waste in office buildings with a fleet of
  mobile manipulators.
\newblock \emph{arXiv preprint arXiv:2305.03270}, 2023.

\bibitem[Hoque et~al.(2025)Hoque, Huang, Yoon, Sivapurapu, and
  Zhang]{hoque2025egodex}
R.~Hoque, P.~Huang, D.~J. Yoon, M.~Sivapurapu, and J.~Zhang.
\newblock Egodex: Learning dexterous manipulation from large-scale egocentric
  video.
\newblock \emph{arXiv preprint arXiv:2505.11709}, 2025.

\bibitem[Hu et~al.(2023{\natexlab{a}})Hu, Rovinsky, Luo, Kumar, Gupta, and
  Levine]{hu2023reboot}
Z.~Hu, A.~Rovinsky, J.~Luo, V.~Kumar, A.~Gupta, and S.~Levine.
\newblock Reboot: Reuse data for bootstrapping efficient real-world dexterous
  manipulation.
\newblock \emph{arXiv preprint arXiv:2309.03322}, 2023{\natexlab{a}}.

\bibitem[Hu et~al.(2023{\natexlab{b}})Hu, Rovinsky, Luo, Kumar, Gupta, and
  Levine]{pmlr-v229-hu23a}
Z.~Hu, A.~Rovinsky, J.~Luo, V.~Kumar, A.~Gupta, and S.~Levine.
\newblock Reboot: Reuse data for bootstrapping efficient real-world dexterous
  manipulation.
\newblock In J.~Tan, M.~Toussaint, and K.~Darvish, editors, \emph{Proceedings
  of The 7th Conference on Robot Learning}, volume 229 of \emph{Proceedings of
  Machine Learning Research}, pages 1930--1949. PMLR, 06--09 Nov
  2023{\natexlab{b}}.
\newblock URL \url{https://proceedings.mlr.press/v229/hu23a.html}.

\bibitem[Intelligence et~al.(2025)Intelligence, Black, Brown, Darpinian,
  Dhabalia, Driess, Esmail, Equi, Finn, Fusai, et~al.]{intelligence2504pi0}
P.~Intelligence, K.~Black, N.~Brown, J.~Darpinian, K.~Dhabalia, D.~Driess,
  A.~Esmail, M.~Equi, C.~Finn, N.~Fusai, et~al.
\newblock $\pi$0. 5: a vision-language-action model with open-world
  generalization, 2025.
\newblock \emph{URL https://arxiv. org/abs/2504.16054}, 1\penalty0
  (2):\penalty0 3, 2025.

\bibitem[Jhamtani and Berg-Kirkpatrick(2018)]{jhamtani2018learning}
H.~Jhamtani and T.~Berg-Kirkpatrick.
\newblock Learning to describe differences between pairs of similar images.
\newblock \emph{arXiv preprint arXiv:1808.10584}, 2018.

\bibitem[Kang et~al.(2025)Kang, Hu, Luo, Yuan, Zheng, and Xu]{kang2025forget}
Z.~Kang, C.~Hu, Y.~Luo, Z.~Yuan, R.~Zheng, and H.~Xu.
\newblock A forget-and-grow strategy for deep reinforcement learning scaling in
  continuous control.
\newblock \emph{arXiv preprint arXiv:2507.02712}, 2025.

\bibitem[Khazatsky et~al.(2024)Khazatsky, Pertsch, Nair, Balakrishna, Dasari,
  Karamcheti, Nasiriany, Srirama, Chen, Ellis, et~al.]{khazatsky2024droid}
A.~Khazatsky, K.~Pertsch, S.~Nair, A.~Balakrishna, S.~Dasari, S.~Karamcheti,
  S.~Nasiriany, M.~K. Srirama, L.~Y. Chen, K.~Ellis, et~al.
\newblock Droid: A large-scale in-the-wild robot manipulation dataset.
\newblock \emph{arXiv preprint arXiv:2403.12945}, 2024.

\bibitem[Kim et~al.(2024)Kim, Pertsch, Karamcheti, Xiao, Balakrishna, Nair,
  Rafailov, Foster, Lam, Sanketi, Vuong, Kollar, Burchfiel, Tedrake, Sadigh,
  Levine, Liang, and Finn]{kim24openvla}
M.~Kim, K.~Pertsch, S.~Karamcheti, T.~Xiao, A.~Balakrishna, S.~Nair,
  R.~Rafailov, E.~Foster, G.~Lam, P.~Sanketi, Q.~Vuong, T.~Kollar,
  B.~Burchfiel, R.~Tedrake, D.~Sadigh, S.~Levine, P.~Liang, and C.~Finn.
\newblock Openvla: An open-source vision-language-action model.
\newblock \emph{arXiv preprint arXiv:2406.09246}, 2024.

\bibitem[Kim et~al.(2025)Kim, Finn, and Liang]{kim2025fine}
M.~J. Kim, C.~Finn, and P.~Liang.
\newblock Fine-tuning vision-language-action models: Optimizing speed and
  success.
\newblock \emph{arXiv preprint arXiv:2502.19645}, 2025.

\bibitem[Kumar et~al.(2024)Kumar, Silver, McClinton, Zhao, Proulx,
  Lozano-Pérez, Kaelbling, and Barry]{kumar2024practice}
N.~Kumar, T.~Silver, W.~McClinton, L.~Zhao, S.~Proulx, T.~Lozano-Pérez, L.~P.
  Kaelbling, and J.~Barry.
\newblock Practice makes perfect: Planning to learn skill parameter policies.
\newblock In \emph{Robotics: Science and Systems (RSS)}, 2024.

\bibitem[Li et~al.(2025)Li, Zhou, and Levine]{li2025reinforcement}
Q.~Li, Z.~Zhou, and S.~Levine.
\newblock Reinforcement learning with action chunking.
\newblock \emph{arXiv preprint arXiv:2507.07969}, 2025.

\bibitem[Lin et~al.(2024)Lin, Hu, Sheng, Wen, You, and Gao]{lin2024data}
F.~Lin, Y.~Hu, P.~Sheng, C.~Wen, J.~You, and Y.~Gao.
\newblock Data scaling laws in imitation learning for robotic manipulation.
\newblock \emph{arXiv preprint arXiv:2410.18647}, 2024.

\bibitem[Liu et~al.(2023)Liu, Li, Wu, and Lee]{liu2023visual}
H.~Liu, C.~Li, Q.~Wu, and Y.~J. Lee.
\newblock Visual instruction tuning.
\newblock \emph{Advances in neural information processing systems},
  36:\penalty0 34892--34916, 2023.

\bibitem[Lu et~al.(2025)Lu, Ye, Ye, Tao, Yang, and Zhao]{lu2025robofac}
W.~Lu, M.~Ye, Z.~Ye, R.~Tao, S.~Yang, and B.~Zhao.
\newblock Robofac: A comprehensive framework for robotic failure analysis and
  correction.
\newblock \emph{arXiv preprint arXiv:2505.12224}, 2025.

\bibitem[Luo et~al.(2024{\natexlab{a}})Luo, Hu, Xu, Tan, Berg, Sharma, Schaal,
  Finn, Gupta, and Levine]{luo2024serl}
J.~Luo, Z.~Hu, C.~Xu, Y.~L. Tan, J.~Berg, A.~Sharma, S.~Schaal, C.~Finn,
  A.~Gupta, and S.~Levine.
\newblock Serl: A software suite for sample-efficient robotic reinforcement
  learning, 2024{\natexlab{a}}.

\bibitem[Luo et~al.(2024{\natexlab{b}})Luo, Xu, Wu, and Levine]{luo2024hilserl}
J.~Luo, C.~Xu, J.~Wu, and S.~Levine.
\newblock Precise and dexterous robotic manipulation via human-in-the-loop
  reinforcement learning, 2024{\natexlab{b}}.

\bibitem[Luo et~al.(2024{\natexlab{c}})Luo, Xu, Wu, and Levine]{luo2024precise}
J.~Luo, C.~Xu, J.~Wu, and S.~Levine.
\newblock Precise and dexterous robotic manipulation via human-in-the-loop
  reinforcement learning.
\newblock \emph{arXiv preprint arXiv:2410.21845}, 2024{\natexlab{c}}.

\bibitem[Luo et~al.(2025)Luo, Xu, Liu, Tan, Lin, Wu, Abbeel, and
  Levine]{luo2025fmb}
J.~Luo, C.~Xu, F.~Liu, L.~Tan, Z.~Lin, J.~Wu, P.~Abbeel, and S.~Levine.
\newblock Fmb: a functional manipulation benchmark for generalizable robotic
  learning.
\newblock \emph{The International Journal of Robotics Research}, 44\penalty0
  (4):\penalty0 592--606, 2025.

\bibitem[Lv et~al.(2025)Lv, Li, Luo, Sun, Kong, Xu, and Ma]{lv2025flow}
L.~Lv, Y.~Li, Y.~Luo, F.~Sun, T.~Kong, J.~Xu, and X.~Ma.
\newblock Flow-based policy for online reinforcement learning.
\newblock \emph{arXiv preprint arXiv:2506.12811}, 2025.

\bibitem[Ma et~al.(2022)Ma, Sodhani, Jayaraman, Bastani, Kumar, and
  Zhang]{ma2022vip}
Y.~J. Ma, S.~Sodhani, D.~Jayaraman, O.~Bastani, V.~Kumar, and A.~Zhang.
\newblock Vip: Towards universal visual reward and representation via
  value-implicit pre-training.
\newblock In \emph{Deep Reinforcement Learning Workshop NeurIPS 2022}, 2022.

\bibitem[Ma et~al.(2023)Ma, Kumar, Zhang, Bastani, and Jayaraman]{ma2023liv}
Y.~J. Ma, V.~Kumar, A.~Zhang, O.~Bastani, and D.~Jayaraman.
\newblock Liv: Language-image representations and rewards for robotic control.
\newblock In \emph{International Conference on Machine Learning}, pages
  23301--23320. PMLR, 2023.

\bibitem[Ma et~al.(2024)Ma, Hejna, Fu, Shah, Liang, Xu, Kirmani, Xu, Driess,
  Xiao, et~al.]{ma2024vision}
Y.~J. Ma, J.~Hejna, C.~Fu, D.~Shah, J.~Liang, Z.~Xu, S.~Kirmani, P.~Xu,
  D.~Driess, T.~Xiao, et~al.
\newblock Vision language models are in-context value learners.
\newblock In \emph{The Thirteenth International Conference on Learning
  Representations}, 2024.

\bibitem[Mendonca et~al.(2023{\natexlab{a}})Mendonca, Bahl, and
  Pathak]{10161016}
R.~Mendonca, S.~Bahl, and D.~Pathak.
\newblock Alan: Autonomously exploring robotic agents in the real world.
\newblock In \emph{2023 IEEE International Conference on Robotics and
  Automation (ICRA)}, pages 3044--3050, 2023{\natexlab{a}}.
\newblock \doi{10.1109/ICRA48891.2023.10161016}.

\bibitem[Mendonca et~al.(2023{\natexlab{b}})Mendonca, Bahl, and
  Pathak]{mendonca2023alan}
R.~Mendonca, S.~Bahl, and D.~Pathak.
\newblock Alan: Autonomously exploring robotic agents in the real world.
\newblock \emph{arXiv preprint arXiv:2302.06604}, 2023{\natexlab{b}}.

\bibitem[Mendonca et~al.(2024)Mendonca, Panov, Bucher, Wang, and
  Pathak]{mendonca2024continuously}
R.~Mendonca, E.~Panov, B.~Bucher, J.~Wang, and D.~Pathak.
\newblock Continuously improving mobile manipulation with autonomous real-world
  rl.
\newblock \emph{arXiv preprint arXiv:2409.20568}, 2024.

\bibitem[Park et~al.(2025)Park, Li, and Levine]{park2025flow}
S.~Park, Q.~Li, and S.~Levine.
\newblock Flow q-learning.
\newblock \emph{arXiv preprint arXiv:2502.02538}, 2025.

\bibitem[Pei et~al.(2025)Pei, Huang, Xu, Chen, He, Yang, Wang, Xie, Qiao, Wu,
  et~al.]{pei2025modeling}
B.~Pei, Y.~Huang, J.~Xu, G.~Chen, Y.~He, L.~Yang, Y.~Wang, W.~Xie, Y.~Qiao,
  F.~Wu, et~al.
\newblock Modeling fine-grained hand-object dynamics for egocentric video
  representation learning.
\newblock \emph{arXiv preprint arXiv:2503.00986}, 2025.

\bibitem[Pertsch et~al.(2025)Pertsch, Stachowicz, Ichter, Driess, Nair, Vuong,
  Mees, Finn, and Levine]{pertsch2025fast}
K.~Pertsch, K.~Stachowicz, B.~Ichter, D.~Driess, S.~Nair, Q.~Vuong, O.~Mees,
  C.~Finn, and S.~Levine.
\newblock Fast: Efficient action tokenization for vision-language-action
  models.
\newblock \emph{arXiv preprint arXiv:2501.09747}, 2025.

\bibitem[Psenka et~al.(2024)Psenka, Escontrela, Abbeel, and Ma]{psenka2024qsm}
M.~Psenka, A.~Escontrela, P.~Abbeel, and Y.~Ma.
\newblock Learning a diffusion model policy from rewards via q-score matching.
\newblock In \emph{Proceedings of the 41st International Conference on Machine
  Learning}, 2024.

\bibitem[Ren et~al.(2024)Ren, Lidard, Ankile, Simeonov, Agrawal, Majumdar,
  Burchfiel, Dai, and Simchowitz]{dppo2024}
A.~Z. Ren, J.~Lidard, L.~L. Ankile, A.~Simeonov, P.~Agrawal, A.~Majumdar,
  B.~Burchfiel, H.~Dai, and M.~Simchowitz.
\newblock Diffusion policy policy optimization.
\newblock In \emph{arXiv preprint arXiv:2409.00588}, 2024.

\bibitem[Schulman et~al.(2017)Schulman, Wolski, Dhariwal, Radford, and
  Klimov]{schulman2017proximal}
J.~Schulman, F.~Wolski, P.~Dhariwal, A.~Radford, and O.~Klimov.
\newblock Proximal policy optimization algorithms.
\newblock \emph{arXiv preprint arXiv:1707.06347}, 2017.

\bibitem[Sermanet et~al.(2024)Sermanet, Ding, Zhao, Xia, Dwibedi,
  Gopalakrishnan, Chan, Dulac-Arnold, Maddineni, Joshi,
  et~al.]{sermanet2024robovqa}
P.~Sermanet, T.~Ding, J.~Zhao, F.~Xia, D.~Dwibedi, K.~Gopalakrishnan, C.~Chan,
  G.~Dulac-Arnold, S.~Maddineni, N.~J. Joshi, et~al.
\newblock Robovqa: Multimodal long-horizon reasoning for robotics.
\newblock In \emph{2024 IEEE International Conference on Robotics and
  Automation (ICRA)}, pages 645--652. IEEE, 2024.

\bibitem[Shafiullah et~al.(2023)Shafiullah, Rai, Etukuru, Liu, Misra, Chintala,
  and Pinto]{shafiullah2023bringing}
N.~M.~M. Shafiullah, A.~Rai, H.~Etukuru, Y.~Liu, I.~Misra, S.~Chintala, and
  L.~Pinto.
\newblock On bringing robots home.
\newblock \emph{arXiv preprint arXiv:2311.16098}, 2023.

\bibitem[Shao et~al.(2024)Shao, Wang, Zhu, Xu, Song, Bi, Zhang, Zhang, Li, Wu,
  et~al.]{shao2024deepseekmath}
Z.~Shao, P.~Wang, Q.~Zhu, R.~Xu, J.~Song, X.~Bi, H.~Zhang, M.~Zhang, Y.~Li,
  Y.~Wu, et~al.
\newblock Deepseekmath: Pushing the limits of mathematical reasoning in open
  language models.
\newblock \emph{arXiv preprint arXiv:2402.03300}, 2024.

\bibitem[Smith et~al.(2023)Smith, Kostrikov, and
  Levine]{smith2023demonstrating}
L.~Smith, I.~Kostrikov, and S.~Levine.
\newblock Demonstrating a walk in the park: Learning to walk in 20 minutes with
  model-free reinforcement learning.
\newblock \emph{Robotics: Science and Systems (RSS) Demo}, 2\penalty0
  (3):\penalty0 4, 2023.

\bibitem[Smith et~al.(2024)Smith, Cao, and Levine]{smith2024grow}
L.~Smith, Y.~Cao, and S.~Levine.
\newblock Grow your limits: Continuous improvement with real-world rl for
  robotic locomotion.
\newblock In \emph{2024 IEEE International Conference on Robotics and
  Automation (ICRA)}, pages 10829--10836. IEEE, 2024.

\bibitem[Team et~al.(2025)Team, Abeyruwan, Ainslie, Alayrac, Arenas, Armstrong,
  Balakrishna, Baruch, Bauza, Blokzijl, et~al.]{team2025gemini}
G.~R. Team, S.~Abeyruwan, J.~Ainslie, J.-B. Alayrac, M.~G. Arenas,
  T.~Armstrong, A.~Balakrishna, R.~Baruch, M.~Bauza, M.~Blokzijl, et~al.
\newblock Gemini robotics: Bringing ai into the physical world.
\newblock \emph{arXiv preprint arXiv:2503.20020}, 2025.

\bibitem[Walke et~al.(2023)Walke, Black, Zhao, Vuong, Zheng, Hansen-Estruch,
  He, Myers, Kim, Du, et~al.]{walke2023bridgedata}
H.~R. Walke, K.~Black, T.~Z. Zhao, Q.~Vuong, C.~Zheng, P.~Hansen-Estruch, A.~W.
  He, V.~Myers, M.~J. Kim, M.~Du, et~al.
\newblock Bridgedata v2: A dataset for robot learning at scale.
\newblock In \emph{Conference on Robot Learning}, pages 1723--1736. PMLR, 2023.

\bibitem[Wang et~al.(2025)Wang, Song, Liu, Ma, Feng, Wang, Jiang, Chen, Zhan,
  Wang, et~al.]{wang2025genie}
W.~Wang, J.~Song, C.~Liu, J.~Ma, S.~Feng, J.~Wang, Y.~Jiang, K.~Chen, S.~Zhan,
  Y.~Wang, et~al.
\newblock Genie centurion: Accelerating scalable real-world robot training with
  human rewind-and-refine guidance.
\newblock \emph{arXiv preprint arXiv:2505.18793}, 2025.

\bibitem[Wang et~al.(2024)Wang, Sun, Zhang, Xian, Biyik, Held, and
  Erickson]{wang2024}
Y.~Wang, Z.~Sun, J.~Zhang, Z.~Xian, E.~Biyik, D.~Held, and Z.~Erickson.
\newblock Rl-vlm-f: Reinforcement learning from vision language foundation
  model feedback.
\newblock In \emph{Proceedings of the 41th International Conference on Machine
  Learning}, 2024.

\bibitem[Xiong et~al.(2024)Xiong, Mendonca, Shaw, and
  Pathak]{xiong2024adaptive}
H.~Xiong, R.~Mendonca, K.~Shaw, and D.~Pathak.
\newblock Adaptive mobile manipulation for articulated objects in the open
  world.
\newblock \emph{arXiv preprint arXiv:2401.14403}, 2024.

\bibitem[Xu et~al.(2022)Xu, Hu, Doshi, Rovinsky, Kumar, Gupta, and
  Levine]{xu2022dexterous}
K.~Xu, Z.~Hu, R.~Doshi, A.~Rovinsky, V.~Kumar, A.~Gupta, and S.~Levine.
\newblock Dexterous manipulation from images: Autonomous real-world rl via
  substep guidance.
\newblock \emph{arXiv preprint arXiv:2212.09902}, 2022.

\bibitem[Yang et~al.(2024)Yang, Tjia, Berg, Damen, Agrawal, and
  Gupta]{10610873}
D.~Yang, D.~Tjia, J.~Berg, D.~Damen, P.~Agrawal, and A.~Gupta.
\newblock Rank2reward: Learning shaped reward functions from passive video.
\newblock In \emph{2024 IEEE International Conference on Robotics and
  Automation (ICRA)}, pages 2806--2813, 2024.
\newblock \doi{10.1109/ICRA57147.2024.10610873}.

\bibitem[Zhai et~al.(2024)Zhai, Wang, Zhang, Huang, Zhang, Zhou, Hou, Qiao, and
  Liu]{zhai2024buildingopenendedembodiedagent}
S.~Zhai, J.~Wang, T.~Zhang, F.~Huang, Q.~Zhang, M.~Zhou, J.~Hou, Y.~Qiao, and
  Y.~Liu.
\newblock Building open-ended embodied agent via language-policy bidirectional
  adaptation, 2024.
\newblock URL \url{https://arxiv.org/abs/2401.00006}.

\bibitem[Zhang et~al.({\natexlab{a}})Zhang, Zhou, Zhai, Sun, and
  Xiong]{zhangefficient}
H.~Zhang, M.~Zhou, S.~Zhai, Y.~Sun, and H.~Xiong.
\newblock Efficient skill discovery via regret-aware optimization.
\newblock In \emph{Forty-second International Conference on Machine Learning},
  {\natexlab{a}}.

\bibitem[Zhang et~al.({\natexlab{b}})Zhang, Zhang, and Gu]{zhangenergy}
S.~Zhang, W.~Zhang, and Q.~Gu.
\newblock Energy-weighted flow matching for offline reinforcement learning.
\newblock In \emph{The Thirteenth International Conference on Learning
  Representations}, {\natexlab{b}}.

\bibitem[Zhou et~al.(2024{\natexlab{a}})Zhou, Atreya, Lee, Walke, Mees, and
  Levine]{zhou2024autonomous}
Z.~Zhou, P.~Atreya, A.~Lee, H.~Walke, O.~Mees, and S.~Levine.
\newblock Autonomous improvement of instruction following skills via foundation
  models.
\newblock \emph{arXiv preprint arXiv:407.20635}, 2024{\natexlab{a}}.

\bibitem[Zhou et~al.(2024{\natexlab{b}})Zhou, Peng, Li, Levine, and
  Kumar]{zhou2024efficient}
Z.~Zhou, A.~Peng, Q.~Li, S.~Levine, and A.~Kumar.
\newblock Efficient online reinforcement learning fine-tuning need not retain
  offline data.
\newblock \emph{arXiv preprint arXiv:2412.07762}, 2024{\natexlab{b}}.

\end{thebibliography}

\appendix
\newpage

\section{Evalauation tasks}
\begin{figure}[ht!]
    \centering
    \includegraphics[width=.95\linewidth]{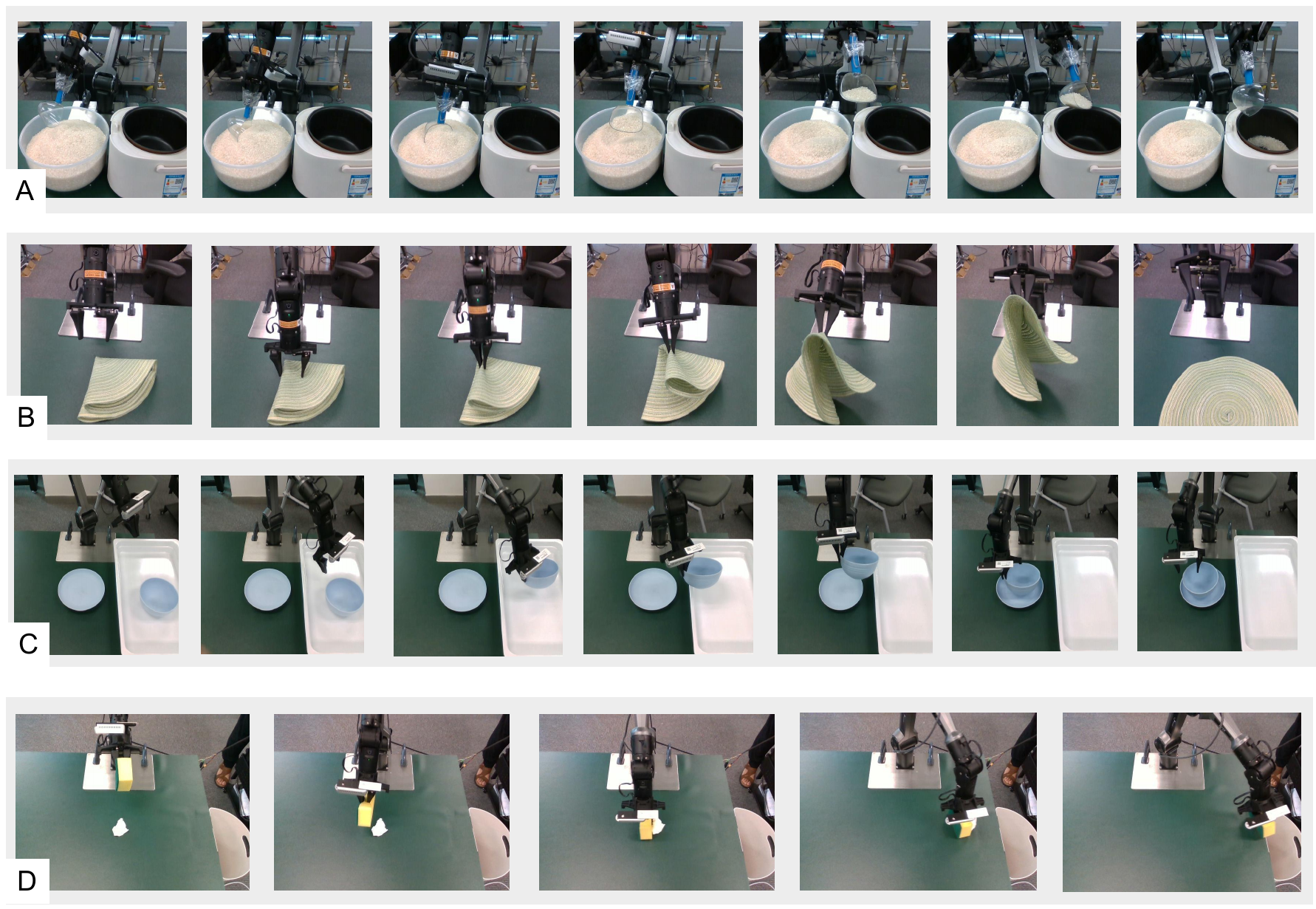}
    \caption{Illustrations of the real-world tasks in our experiments. (A) Rice Scooping and Transfer: scoop the rice from the big bowl and transfer it to the cooker. (B) Unfold Mat: unford the folded mat on the table. (C) Pick and Place Bowl: pick up the bowl from the tray and place it on the plate. (D) Desktop Sweep Disposal:  sweep the trash on the table into the trash can.} 
    \label{fig:task_info}
\end{figure}

We design five manipulation tasks in the real-world RL experiments as shown in Figure~\ref{fig:task_info}.
These tasks are expected to constitute a rich and relatively complete kitchen scene from cooking food to setting up the table. 
In addition, these manipulation tasks are diverse from two perspectives: 1) manipulating different types of objects, i.e., rigid objects (C), flexible objects (B,D), and granular objects (A); 2) requires different manipulation ability, either precise goal reaching, i.e., touch/push objects (D), or grasp proper parts (B,C), or dynamic manipulation (A). 


\textbf{Rice Scooping and Transfer} In this task, the robot is assumed to firstly scoop a spoonful of rice from the jar, and then pour the rice into the cooker. The task is considered successful if the rice is transferred from the jar to cooker without spilling. The difficulty lies in the granular objects are not easy to be obtained and the transportation also requires the robot to be very stable. This task is also named \textbf{Rice Transfer} for abbreviation.

\textbf{Unfold Mat} In this task, a mat is folded on the desktop, and the robot is supposed to grab the mat, lift it up, and then release it to unfold this mat. This task is considered successful if the mat unfolds well on the table. The difficulty lies in two points: 1) the robot must grab a proper part of the mat, the bottom, middle and far side of the mat is hard to grad or unfold; 2) the mat must be raised high enough so that it can fall and spread naturally, otherwise it may still be folded after falling on the table.

\textbf{Pick and Place Bowl} In this task, a plate is placed on the table and a bowl is placed in a tray, and the robot is supposed to pick up the bowl and place it on the plate. The task is considered successful if the bowl is properly put on the plate. This task requires precise and gentle gripping of the bowl's rim and delivering it precisely to the center of the plate.

\textbf{Desktop Sweep Disposal} In this task, there is a white trash on the table and the robot is supposed to sweep it into the trash can. The task is considered successful if the trash is swept into the nearby can. Similar to task (A), this task requires the robot to precisely reach the trash and push it with a proper force. Too much or too little force can lead to failure. This task is also named \textbf{Desktop Sweep} for abbreviation.

\section{Dataset}
\label{sec:dataset}
\hspace{1.5em}LLAVA~\cite{liu2023visual}: This dataset includes basic multi-turn dialogues and a rich collection of VQA data, which helps maintain the model’s general multimodal understanding and conversational abilities.

RoboVQA~\cite{sermanet2024robovqa}: This dataset contains VQA data from various robotic tasks, aimed at improving the model's understanding of multimodal data in robotic scenarios.

SpatialVQA~\cite{chen2024spatialvlm}: This dataset provides spatial reasoning data based on RGB images, including depth estimation, object detection, and more, which strengthens the model's ability to understand and estimate spatial information within images.

Spot-the-diff~\cite{jhamtani2018learning}, InstructPix2Pix~\cite{brooks2023instructpix2pix}: These datasets consist of pair-wise image difference comparison data, designed to enhance the model’s capability to detect fine-grained differences between images, thereby supporting pair-wise progress understanding tasks.
\begin{table}[htbp]
\centering
\renewcommand{\arraystretch}{1.1}
\adjustbox{width=\textwidth,center}
{
\begin{tabular}{|l|l|l|l|l|}
\hline
\textbf{Dataset Name} & \textbf{Size/samples} & \textbf{Size/h} & \textbf{Tasks} & \textbf{Mixture Weight} \\ \hline
Ego4D HOD~\cite{pei2025modeling}            & 157M                  & 3k              & TPU            & 14.6\%                  \\ \hline
InstructPix2Pix~\cite{brooks2023instructpix2pix}       & 36k                   & -               & GVL            & 0.4\%                   \\ \hline
RobotVQA~\cite{sermanet2024robovqa}              & 50k                   & -               & GVL            & 0.6\%                   \\ \hline
Spot the diff~\cite{jhamtani2018learning}         & 27k                   & -               & GVL            & 0.3\%                   \\ \hline
Llava~\cite{liu2023visual}                 & 633k                  & -               & GVL            & 4.7\%                   \\ \hline
SpatialQA~\cite{chen2024spatialvlm}             & 781k                  & -               & GVL            & 5.8\%                   \\ \hline
AGIBOT~\cite{bu2025agibot}               & 8M                    & 73              & TPU,GVL,VLA    & 3.0\%                   \\ \hline
Bridge~\cite{walke2023bridgedata}               & 2M                    & 135             & TPU,VLA        & 9.1\%                   \\ \hline
Droid~\cite{khazatsky2024droid}                 & 40M                   & 741             & TPU,VLA        & 30.0\%                  \\ \hline
FMB~\cite{luo2025fmb}                   & 2m                    & 144             & TPU,VLA        & 9.7\%                   \\ \hline
RoboSet~\cite{bharadhwaj2024roboagent}               & 4m                    & 130             & TPU,VLA        & 17.4\%                  \\ \hline
Self Collected        & 946k                  & 18              & TPU,VLA        & 4.4\%                   \\ \hline
\end{tabular}
}
\vspace{-0.1cm}
\footnotesize
\textbf{Note:} TPU = Task Progress Understanding; GVL = General Vision-Language; VLA = Vision-Language-Action.
\caption{Overview of Data Mixture.}
\label{tab:dataset}
\end{table}

\vspace{-0.2cm}
\section{Scene Transfer}
\begin{figure}[H]
    \centering
    \includegraphics[width=0.9\linewidth]{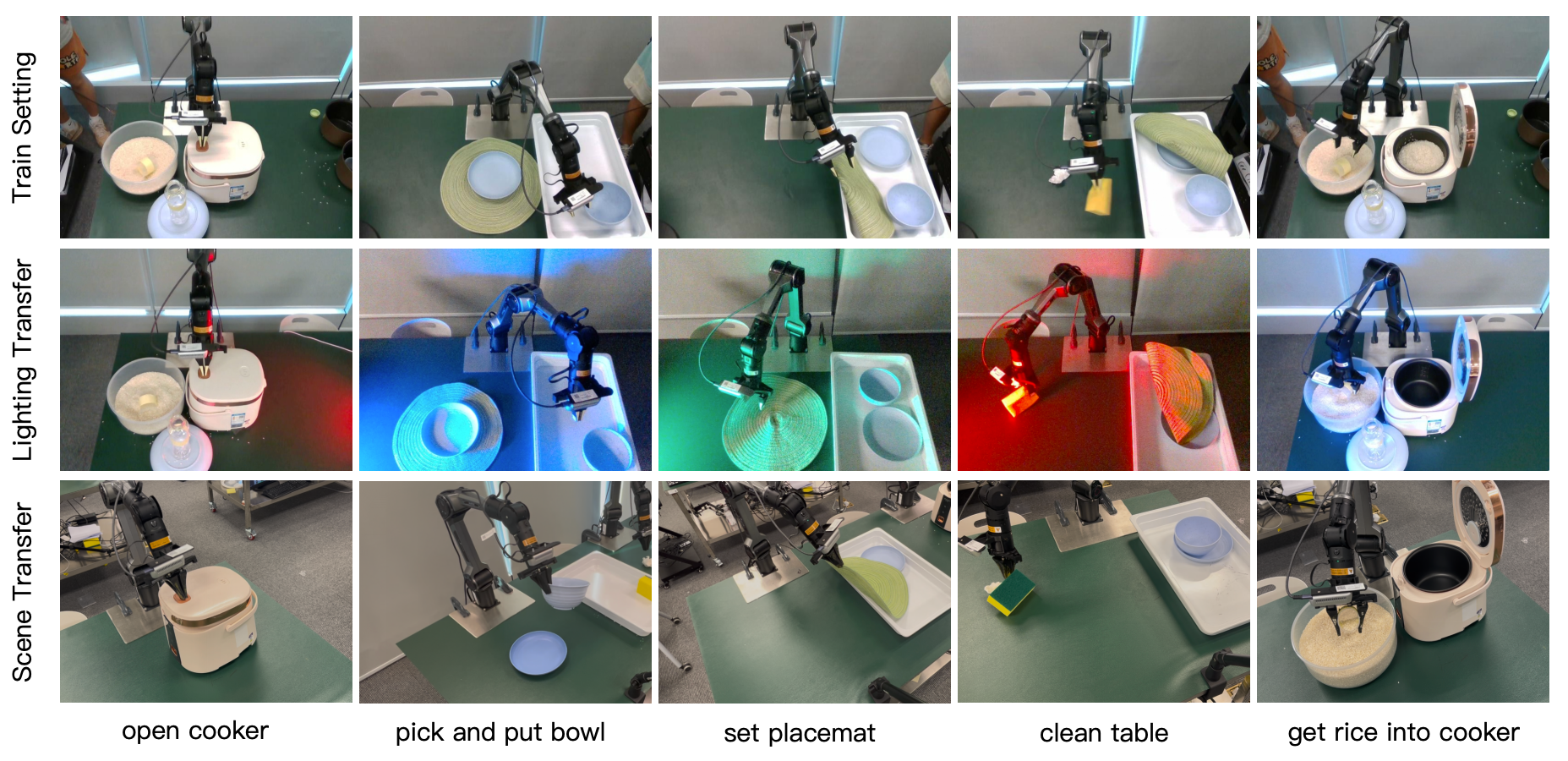}
    \caption{Illustrations of scene/lighting transfer.} %
    \label{fig:transfer}
\end{figure}


\section{Multi-Robots Scaling}
\label{sec:Multi-Robots Scaling}
\begin{figure}[ht] 
  \centering
  \begin{subfigure}{0.3\textwidth} 
    \includegraphics[width=\linewidth]{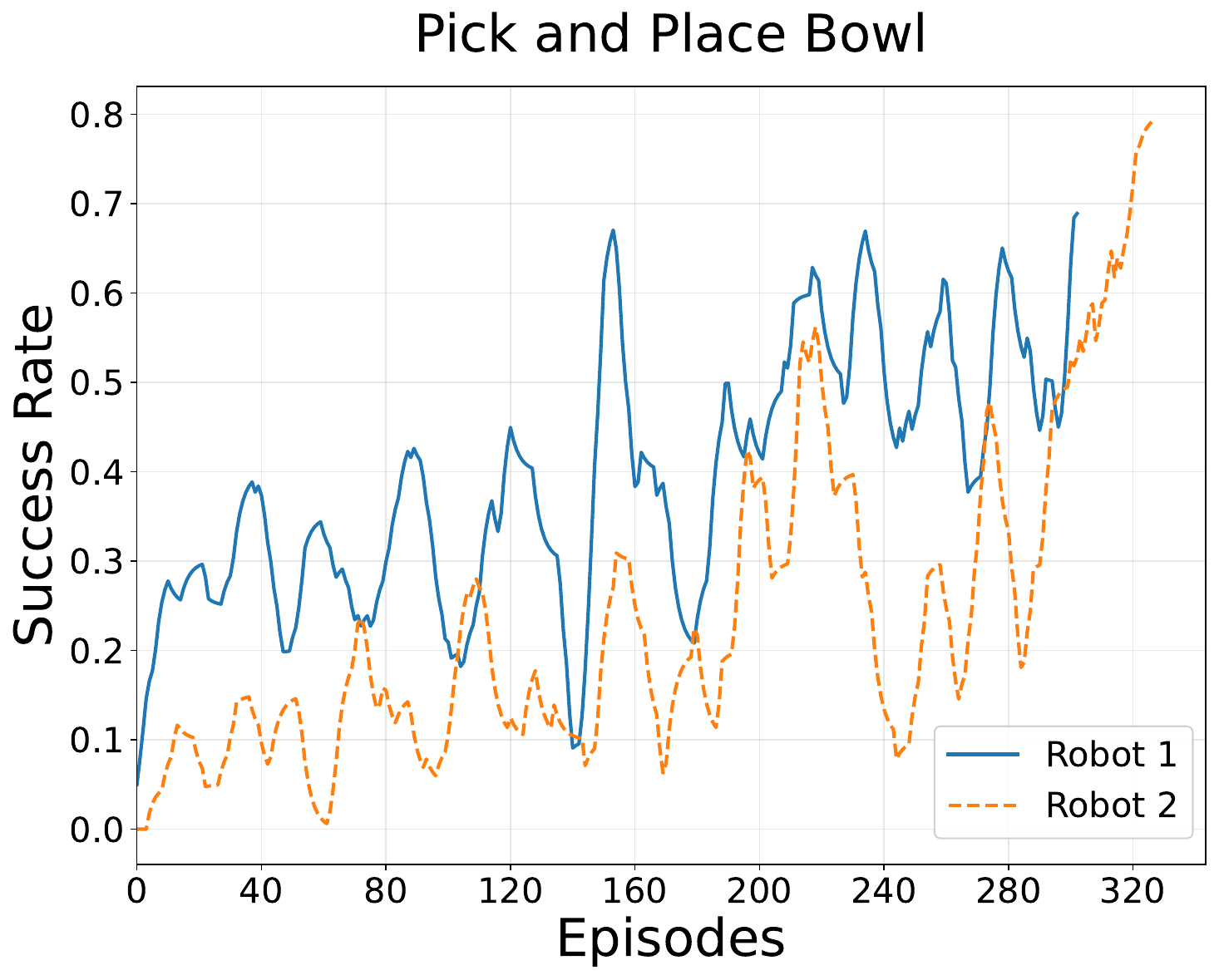}
    \caption{Training with 2 robots} 
    \label{fig:scale_time_sub}
  \end{subfigure}
  \hfill 
  \begin{subfigure}{0.3\textwidth} 
    \includegraphics[width=\linewidth]{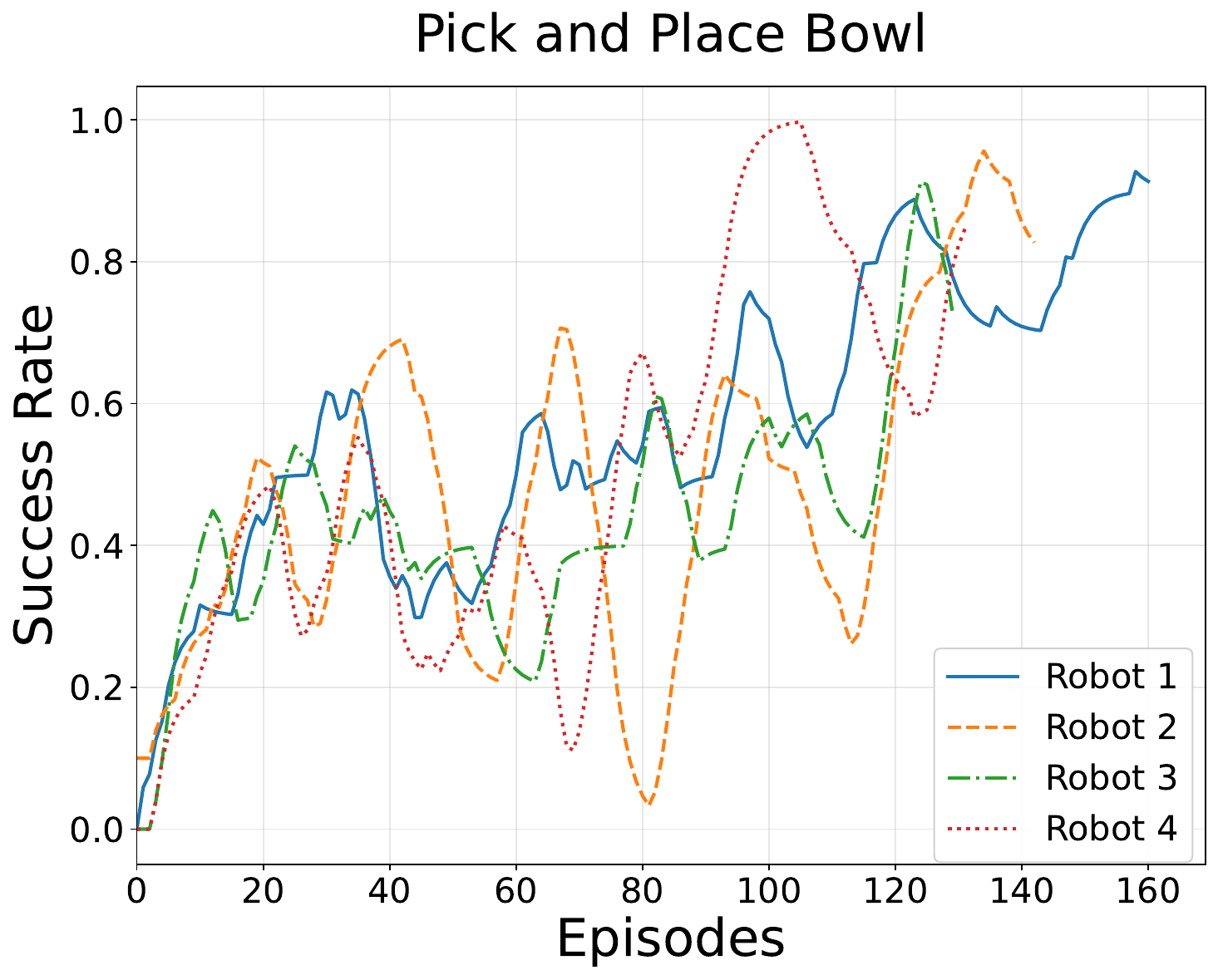}
    \caption{Training with 4 robots} 
    \label{fig:scale_episode_sub}
  \end{subfigure}
  \hfill 
  \begin{subfigure}{0.3\textwidth} 
    \includegraphics[width=\linewidth]{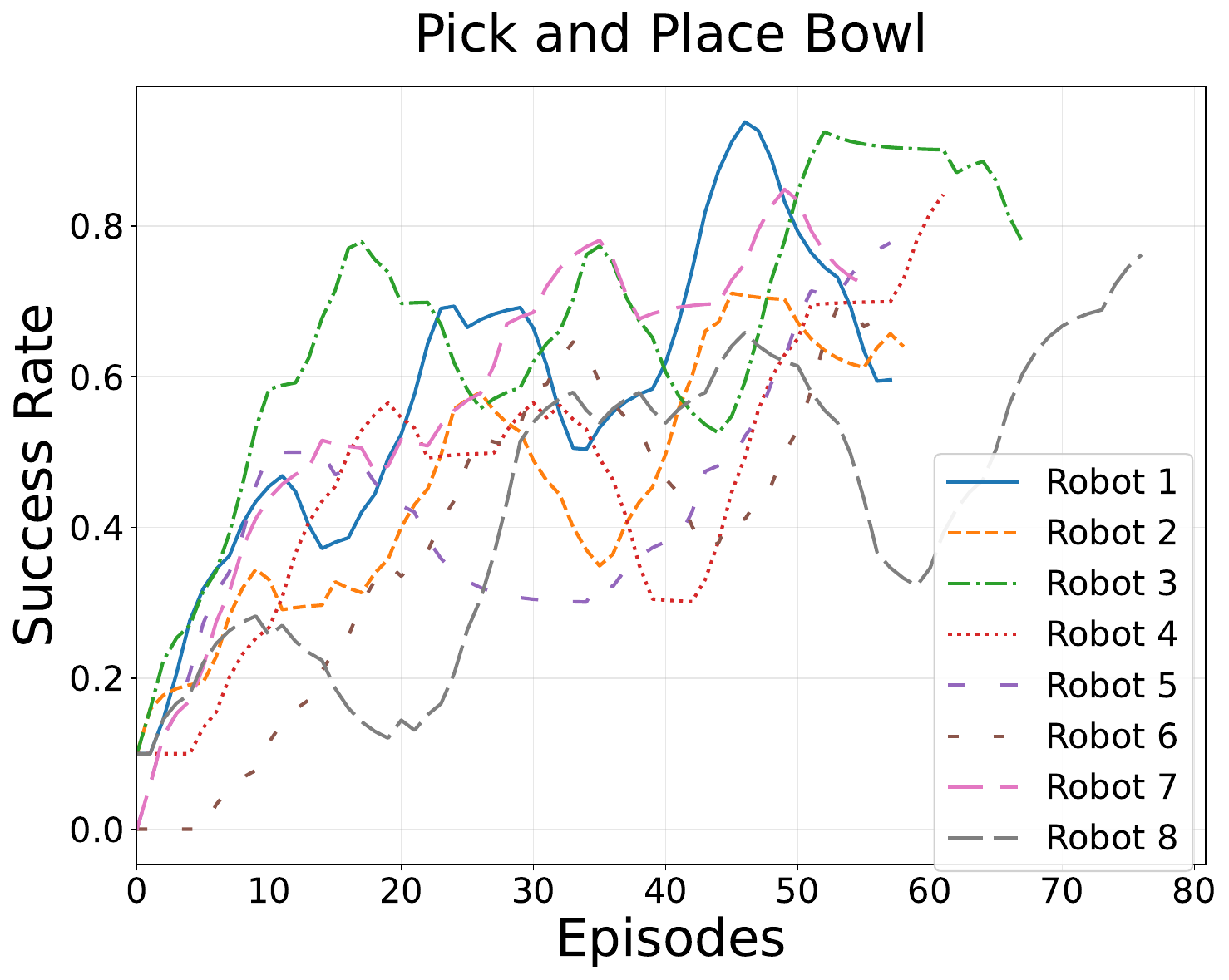} 
    \caption{Training with 8 robots} 
    \label{fig:scale_metric_sub}
  \end{subfigure}
  \caption{Demonstration of the experimental results on Pick and Place Bowl with different amount of robots trained at the same time.} 
  \label{fig:scale_different_robots}
\end{figure}

Real-world RL across multiple robots exposes a learning imbalance driven by differences in background clutter, lighting, and robot status (e.g., temperature). These small covariate shifts lead to divergent per-robot success curves under a shared policy. Robots that have higher success rates tend to learn faster than others, and robots that have lower success rates hardly improve over time. To achieve stable multi-robot training, we adopt a dynamic sampling strategy that increases the sampling frequency of data generated by under-learned robots. In this way, multiple robots can be trained more stably, as shown in Figure \ref{fig:scale_different_robots}.

\section{Author contributions}
\label{sec:authors}
All authors contributed to writing.
\begin{itemize}[leftmargin=*]
    \item Shaopeng Zhai (zhaishaopeng@pjlab.org.cn): Team leadership, led design and development, RL training, infrastructure engineering, robot/data engineering, VLAC pretraining, architecture/algorithm design, experiments analysis, paper writing.
    \item Qi Zhang (zhangqi1@pjlab.org.cn): Robot/data engineering, architecture/algorithm design, experiments analysis, paper writing, and significant contribution to VLAC pretraining.
    \item Tianyi Zhang (zhangtianyi@pjlab.org.cn): Experiments analysis, robot engineering, infrastructure engineering, experiments analysis, and significant contribution to RL training.
    \item Fuxian Huang (huangfuxian@pjlab.org.cn): Experiments analysis, paper writing, and significant contribution to RL training.
    \item Haoran Zhang (zhanghaoran@pjlab.org.cn): Experiments analysis, robot engineering, experiments analysis, and significant contribution to multi-robot RL training.
    \item Ming Zhou (zhouming@pjlab.org.cn): algorithm design and significant contribution to infrastructure engineering.
    \item Shengzhe Zhang (owen\_zsz@mail.ustc.edu.cn): data collection, model testing.
    \item Litao Liu (litao.liu@rutgers.edu): data collection, model testing.
    \item Sixu Lin (linsixu@pjlab.org.cn): data collection, model testing.
    \item Jiangmiao Pang (pangjiangmiao@pjlab.org.cn): Team leadership and advised project direction with critical feedback.
\end{itemize}

\end{document}